\documentclass[letterpaper, 10 pt, conference]{ieeeconf}  

\IEEEoverridecommandlockouts                              

\usepackage{amsmath} 
\usepackage{amssymb}  
\usepackage{bm}  
\usepackage{graphics} 
\usepackage{epsfig} 
\usepackage{wrapfig}
\usepackage{subcaption}
\usepackage[font=small,labelfont=small]{caption}
\usepackage{hyperref}
\usepackage{color}
\usepackage[table]{xcolor}
\usepackage[ruled, linesnumbered]{algorithm2e}
\usepackage{sidecap}
\usepackage{tikz}
\usetikzlibrary{arrows.meta}

\usepackage[per-mode=fraction]{siunitx}
\usepackage{comment}
\usepackage[normalem]{ulem}
\usepackage{xspace}
\usepackage{xcolor}

\DeclareSIUnit\px{px}
\DeclareSIUnit\fps{fps}

\definecolor{OliveGreen}{RGB}{0,200,25}
\newcommand{\red}[1]{\textcolor{red}{#1}}

\newcommand{\darkgreen}[1]{\textcolor{OliveGreen}{#1}}

\newif\iffinal

\newcommand{\replaced}[2]{%
	\iffinal%
	#2%
	\else%
	\red{\ifmmode\text{\sout{\ensuremath{#1}}}\else\sout{#1}\fi}\darkgreen{#2}%
	\fi%
}
\newcommand{\removed}[1]{%
	\iffinal%
	\else%
	\red{\ifmmode\text{\sout{\ensuremath{#1}}}\else\sout{#1}\fi}%
	\fi%
}

\graphicspath{{Figures/}}

\title{\LARGE \bf
Learning to Sequence and Blend Robot Skills \\via Differentiable Optimization
}

\author{No\'emie Jaquier, You Zhou, Julia Starke, and Tamim Asfour
\thanks{This work was supported by the Helmholtz AI project LearnGraspPhases and the Carl Zeiss Foundation through the JuBot project. The authors are with the Institute for Anthropomatics and Robotics, Karlsruhe Institute of Technology, Karlsruhe, Germany. Correspondence to: {\tt\footnotesize \{noemie.jaquier, asfour\}@kit.edu}}%
}

\newcommand{\etal}{\MakeLowercase{\textit{et al.}}}
\newcommand{\trsp}{\mathsf{T}}
\newcommand{\ty}[1]{{\scriptscriptstyle{\mathcal{#1}}}}

\definecolor{darkyellow}{rgb}{0.86, 0.66, 0.}
\definecolor{darkred}{rgb}{0.66, 0.08, 0.23}
\definecolor{darkorange}{rgb}{1., 0.55, 0.0}
\definecolor{orange}{rgb}{1., 0.65, 0.1}
\definecolor{skyblue}{rgb}{0., 0.65, 0.9}
\definecolor{steelblue}{rgb}{0.275, 0.501, 0.706}
\definecolor{gray}{rgb}{0.7, 0.7, 0.7}
\definecolor{kitgreen}{rgb}{0.0, 0.588, 0.51}
\definecolor{kitblue}{rgb}{0.274, 0.392, 0.51}
\definecolor{kitlightgreen}{rgb}{0.540, 0.71, 0.235}

\DeclareRobustCommand{\pickcircle}{\tikz{ \filldraw[color=white, fill=darkyellow!60, thick](0,0) circle (.1);}}
\DeclareRobustCommand{\placecircle}{\tikz{ \filldraw[color=white, fill=darkred!60, thick](0,0) circle (.1);}}
\DeclareRobustCommand{\graytraining}{\raisebox{2pt}{\tikz{\draw[gray,solid,line width = 1.5pt](0,0) -- (3mm,0);}}}
\DeclareRobustCommand{\blackrepro}{\raisebox{2pt}{\tikz{\draw[black,densely dashdotted,line width = 1.1pt](0,0) -- (3mm,0);}}}
\DeclareRobustCommand{\redrepro}{\raisebox{2pt}{\tikz{\draw[darkred,solid,line width = 1.5pt](0,0) -- (3mm,0);}}}
\DeclareRobustCommand{\orangerepro}{\raisebox{2pt}{\tikz{\draw[darkorange,solid,line width = 1.5pt](0,0) -- (3mm,0);}}}
\DeclareRobustCommand{\bluerepro}{\raisebox{2pt}{\tikz{\draw[steelblue,solid,line width = 1.5pt](0,0) -- (3mm,0);}}}
\DeclareRobustCommand{\blueDS}{\raisebox{2pt}{\tikz{\draw[-{Latex[length=1.5mm]}, skyblue,solid,line width = 1.1pt](0,0) -- (3mm,0);}}}

\DeclareRobustCommand{\kitgreenrepro}{\raisebox{2pt}{\tikz{\draw[kitgreen,solid,line width = 1.5pt](0,0) -- (3mm,0);}}}
\DeclareRobustCommand{\kitbluerepro}{\raisebox{2pt}{\tikz{\draw[kitblue,solid,line width = 1.5pt](0,0) -- (3mm,0);}}}
\DeclareRobustCommand{\kitlightgreenrepro}{\raisebox{2pt}{\tikz{\draw[kitlightgreen,solid,line width = 1.5pt](0,0) -- (3mm,0);}}}
\DeclareRobustCommand{\kitorangerepro}{\raisebox{2pt}{\tikz{\draw[orange,solid,line width = 1.5pt](0,0) -- (3mm,0);}}}

\begin{document}

\maketitle
\thispagestyle{empty}
\pagestyle{empty}

\begin{abstract}

In contrast to humans and animals who naturally execute seamless motions, learning and smoothly executing sequences of actions remains a challenge in robotics.
This paper introduces a novel skill-agnostic framework that learns to sequence and blend skills based on differentiable optimization. 
Our approach encodes sequences of previously-defined skills as quadratic programs (QP), whose parameters determine the relative importance of skills along the task. Seamless skill sequences are then learned from demonstrations by exploiting differentiable optimization layers and a tailored loss formulated from the QP optimality conditions. Via the use of differentiable optimization, our work offers novel perspectives on multitask control.
We validate our approach in a pick-and-place scenario with planar robots, a pouring experiment with a real humanoid robot, and a bimanual sweeping task with a human model.
\end{abstract}

\section{Introduction}
\label{sec:Introduction}
Humans and animals generally achieve seamless sequences of actions, featuring smooth and natural transitions. Indeed, there are biological evidences that motor actions are composed of fundamental building blocks, which are then smoothly sequenced and combined to realize complex motions~\cite{MussaIvaldi00:MotorPrimitives, Flash05:MotorPrimitives}. This particularly applies to manipulation tasks, which can be broken down into several smoothly-linked action phases for which the brain selects and executes appropriate controllers~\cite{Johansson09:GraspPhases}. In contrast, learning and executing seamless sequences of actions is still a challenge in robotics. Indeed, skills are usually learned for a specific task and are thus difficult to re-use in a different sequence of actions. Moreover, robot motions are characterized by obvious jerky transitions, which are so typical that people imitate robots by introducing abrupt pauses between subsequent movements. 

In this paper, we propose a novel skill-agnostic approach to sequence and blend skills. To do so, we encode sequences of skills as quadratic programs (QP)~\cite{Nocedal06:OptimizationBook} and leverage differentiable optimization (Optnet) layers~\cite{Amos17:Optnet,Agrawal19:DifferentiableOptLayers} to determine the relative importance of each skill throughout the task (see \S~\ref{sec:Background} for a background). Our approach is skill-agnostic by acting on a set of control values, thus considering skills as a-priori given black-box solutions. Given a set of previously-defined (i.e., learned or programmed) skills and few demonstrations of a task, our formulation not only learns a suitable sequence of possibly-concurrent skills, but also blends transitions "for free", i.e., requiring no additional operations (see \S~\ref{sec:LearningApproach}).

The contributions of this paper are: (\emph{i}) We propose a novel QP-based approach to learn seamless sequences of skills from demonstrations; (\emph{ii}) We formulate a tailored loss function from the optimality of the QP; (\emph{iii}) We present two types of QP parameters to encode the importance of skills; (\emph{iv}) We bring a novel perspective on multitask control via the use of differentiable optimization. We showcase our approach in various experiments with simulated and real robots (\S~\ref{sec:Experiments}).


\section{Related Work}
\label{sec:RelatedWork}
Given a set of individual robotic skills, the challenge is to order and combine them to successfully execute complex manipulation tasks. Sequencing approaches presented in the literature are mainly based on learning from demonstrations (LfD)~\cite{Manschitz15:LearningSequentialSkills,Manschitz15:ConcurrentSequentialSkills,Rozo20:LearningSequencing,Konidaris12:SkillTrees} or on reinforcement learning (RL)~\cite{Konidaris12:SkillTrees, Stulp12:RLSequencingDMP}. 
Manschitz \etal~\cite{Manschitz15:LearningSequentialSkills} learn both a sequence graph of skills from demonstrations, and a classifier to select the transitions. The authors extend their approach to handle concurrent skill activations~\cite{Manschitz15:ConcurrentSequentialSkills}. 
As opposed to our work, the transitions between skills are explicitly labeled for the demonstrations.
Rozo \etal~\cite{Rozo20:LearningSequencing} introduce an object-centered skill sequencing formulation, which builds a complete model of the task by cascading several skill models, and adapting their task parameters. In contrast to our approach, the desired skill sequence is assumed to be given. 
In \cite{Konidaris12:SkillTrees}, demonstrated trajectories are segmented into sequences of skills, where skill policies are represented by linear value function approximations. Sequences from several demonstrations are then combined into skill trees.
Stulp \etal~\cite{Stulp12:RLSequencingDMP} extend the PI$^2$ algorithm to optimize sequences of dynamical movement primitives (DMP) by simultaneously learning their shape and goal parameters.
Overall, the aforementioned approaches are specifically tailored to a single skill type, e.g., dynamical systems~\cite{Manschitz15:LearningSequentialSkills,Manschitz15:ConcurrentSequentialSkills}, task-parametrized Gaussian mixture model (TP-GMM)~\cite{Rozo20:LearningSequencing}, or DMP~\cite{Stulp12:RLSequencingDMP}. Moreover, transitions are usually handled by matching the end- and start-points of subsequent skills, and are thus characterized by obvious pauses.
In contrast, our approach is \emph{skill-agnostic} and learns sequences featuring \emph{seamless and natural} transitions.

Other works focus on designing smooth transitions between skills. For instance, several approaches were presented in~\cite{Saveriano19:BlendingDMP} to blend 
DMPs, and probabilistic movement primitives (ProMP) can naturally be blended~\cite{Paraschos18:ProMP}. However, these methods require a known sequence of specific skills and a manual tuning of transition parameters. 
In~\cite{Luksch12:HierarchicalSequenceMPs}, motions are generated from a hierarchy of motion primitives, which are activated based on a neural-like dynamics. Therefore, sequencing and blending is achieved by choosing suitable weights and connections. This approach was then combined with optimal control for continuous motion adaptation~\cite{Muehlig14:HierarchicalSequenceMPs}. Although it generates seamless motions, its applicability is limited due to the necessity of defining the model by hand.

Sequencing and blending of tasks has also been explored in the context of robot multitask control. Salini \etal~\cite{Salini11:Sequencing} combine different controllers in a QP formulation by defining a soft hierarchy of tasks. This corresponds to defining a sequence of skills with concurrent activations. Smooth transitions are achieved by smoothly-varying the relative importance of skills (priorities) with manually-tuned weights. In~\cite{Dehio15:SoftPriorities}, the skills priorities are instead optimized using covariance matrix adaptation evolution strategy (CMA-ES) in order to superpose several controllers for motion generation. Modugno \etal~\cite{Modugno16:CMAESSoftPriorities} extended this idea to learn time-varying skill priorities given as a weighted sum of basis functions equally spaced in time. The corresponding weights can then be optimized using black-box optimization techniques such as CMA-ES~\cite{Modugno16:CMAESSoftPriorities} or Bayesian optimization (BO)~\cite{Su18:BOTaskPriorities,Li20:BOSoftPriorities}. 
Our work distinguishes in that we directly learn the relative importance of skills along the task by \emph{differentiating} through the optimization problem. In contrast to~\cite{Modugno16:CMAESSoftPriorities,Su18:BOTaskPriorities,Li20:BOSoftPriorities}, we leverage LfD to learn sequences of previously-defined skills with seamless transitions. Therefore, our approach requires only few initial demonstrations and no additional trials during the learning phase, thus improving on data-efficiency and training cost compared to black-box optimization techniques.


\section{Background}
\label{sec:Background}
\subsection{Multitask control with quadratic programming}
\label{subsec:MultitaskControl}
Quadratic programs (QP)~\cite[Chap.\ 16]{Nocedal06:OptimizationBook} are extensively used to formulate multitask control of humanoid robots as a constrained optimization problem. Indeed, QP can be solved very efficiently, while explicitly incorporating a wide variety of objectives and accounting for diverse constraints (see e.g.,~\cite{Bouyarmane16:WeightPriorizedControl, Collette08:QP}). A QP solves a problem of the form
\begin{equation}
\min_{\bm{z}} \tfrac{1}{2} \bm{z}^\trsp \bm{Q} \bm{z} + \bm{c}^\trsp \bm{z} \; \text{ s. t. }\; \bm{A}\bm{z} = \bm{b} \text{ and } \bm{G}\bm{z} \leq \bm{h},
\label{Eq:QP}
\end{equation} 
where $\bm{z}\in\mathbb{R}^n$ is the optimization variable, $\bm{Q}\in\mathcal{S}_{\ty{+}}^n$, $\bm{c}\in\mathbb{R}^n$ are the parameters of the quadratic cost function with $\mathcal{S}_{\ty{+}}^n$ denoting the manifold of positive-semidefinite (PSD) matrices, and $\bm{A}\in\mathbb{R}^{m\times n}$, $\bm{b}\in\mathbb{R}^m$, $\bm{G}\in\mathbb{R}^{p\times n}$, $\bm{h}\in\mathbb{R}^p$ are the constraints parameters.
For robot multitask control, QP are typically used to minimize the weighted sum of a set of $K$ tasks, i.e., $\min_{\bm{\xi}_1 \ldots \bm{\xi}_K} \sum_{k=1}^{K} w_k \|\hat{\bm{\xi}}_k - \bm{\xi}_k \|^2$, where $\hat{\bm{\xi}}_k$ and $\bm{\xi}_k$ are the desired and current value of the task $k$, respectively, and $w_k$ is a weight setting the relative importance of the task $k$ with respect to the other tasks.
Moreover, the constraints typically include the equations of motion (kinematics, or dynamics), the technological limits of the system (e.g., joint limits), and interaction constraints (e.g., grasp or frictional contacts).
In this paper, we use a QP to encode a sequence of skills, along which the weights scaling the importance of each skill vary, leading to smooth trajectories and transitions.

\subsection{Karush-Kuhn-Tucker conditions}
\label{subsec:KKT}
The Karush-Kuhn-Tucker (KKT) conditions~\cite{Kuhn51:KKTconditions} are first order necessary conditions for $\bm{z}^*$ to be a local solution of a constrained optimization problem. In particular, the KKT conditions corresponding to the QP~\eqref{Eq:QP} are (\emph{i}) $\nabla_{\bm{z}} \mathcal{L}(\bm{z}, \bm{\mu}, \bm{\nu}) = 0$ with $\mathcal{L}(\bm{z}, \bm{\mu}, \bm{\nu})$ the Lagrangian function of the problem~\eqref{Eq:QP}, and $\bm{\mu}, \bm{\nu}$ the Lagrangian multipliers corresponding to its equality and inequality constraints, respectively, (\emph{ii}) $ \bm{A}\bm{z} = \bm{b}$, (\emph{iii}) $\bm{G}\bm{z} \leq \bm{h}$, (\emph{iv}) $\mu_i \geq 0 \;\forall i\in\{1 \ldots m \}$, and (\emph{v}) $\nu_j \geq 0 \; \forall j\in\{1 \ldots p \}$.

In addition to being used throughout the solving process of constrained optimization problems, the KKT conditions were exploited in inverse optimal control (IOC). In IOC, trajectories are viewed as the solution of an optimization problem, which aims at minimizing an unknown (parametric) cost. In this context, Englert \etal~\cite{Englert17:InverseKKT} used the fact that demonstrations of such trajectories --- under the assumption that they are optimal --- fulfill the KKT conditions, to determine the optimal parameters of the underlying cost.\footnote{Similar ideas have also been explored in the context of inverse reinforcement learning (IRL), where the parameters of a reward function were selected by minimizing the norm of the expert's policy gradient~\cite{Pirotta16:InverseRL}.} We follow a similar reasoning and leverage the QP KKT conditions to define the loss of our sequencing approach.

\subsection{Differentiable optimization layers}
\label{subsec:DifferentiableOptimization}
Recent works~\cite{Amos17:Optnet, Agrawal19:DifferentiableOptLayers} proposed to integrate optimization layers into neural architectures by differentiating through the corresponding optimization problems. In particular, Amos and Kolter~\cite{Amos17:Optnet} introduced \emph{Optnet}, a neural architecture embedding QP as individual layers. Namely, Optnet defines the output $\bm{z}_{i+1}$ of the current layer as the solution of a QP whose 
parameters depend on the previous layer $\bm{z}_i$, i.e., 
\begin{align}
\bm{z}_{i+1} &= 
\min_{\bm{z}} \tfrac{1}{2} \bm{z}^\trsp \bm{Q}(\bm{z}_i) \bm{z} + \bm{c}(\bm{z}_i)^\trsp \bm{z} \; \nonumber \\ &\text{ s. t. }\; \bm{A}(\bm{z}_i)\bm{z} = \bm{b}(\bm{z}_i) \text{ and } \bm{G}(\bm{z}_i)\bm{z} \leq \bm{h}(\bm{z}_i).
\label{Eq:Optnet}
\end{align} 
In order to train Optnet using backpropagation, the layer~\eqref{Eq:Optnet} must be differentiable, i.e., the derivatives of the solution $\bm{z}_{i+1}$ of the QP with respect to its input parameters $\{\bm{Q},\bm{c},\bm{A},\bm{b},\bm{G}, \bm{h}\}(\bm{z}_i)$ must be computed. This is achieved by differentiating the KKT conditions of the problem at a given solution (see~\cite{Amos17:Optnet}). In this paper, we leverage Optnet to learn the importance of individual skills throughout the task.


\section{Learning to Sequence and Blend Skills}
\label{sec:LearningApproach}
\begin{figure*}[tbp]
	\centering
	\begin{subfigure}[b]{0.11\textwidth}
		\includegraphics[width=\textwidth]{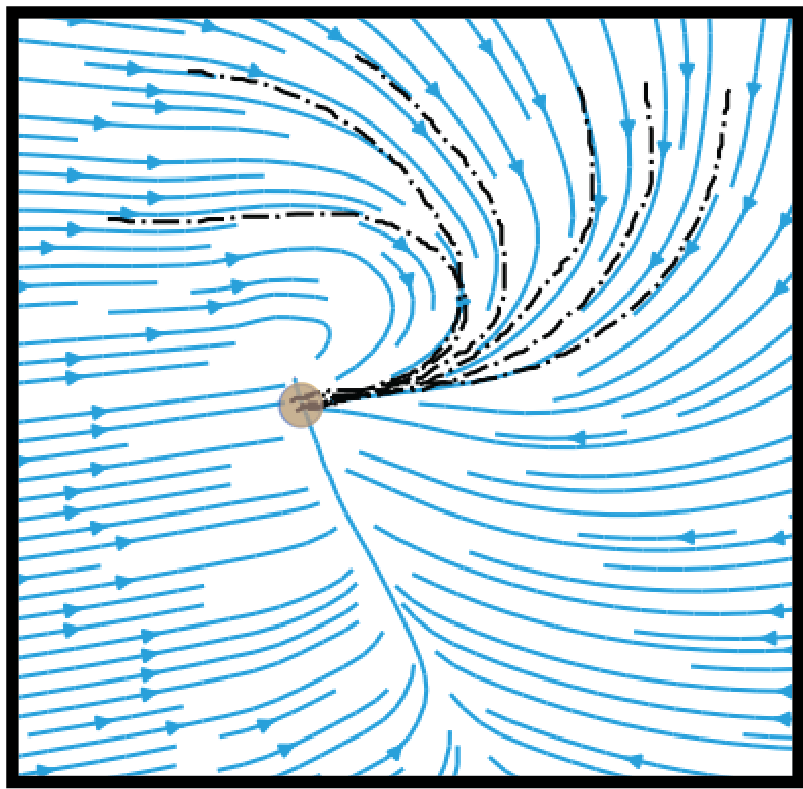}
		\includegraphics[width=\textwidth]{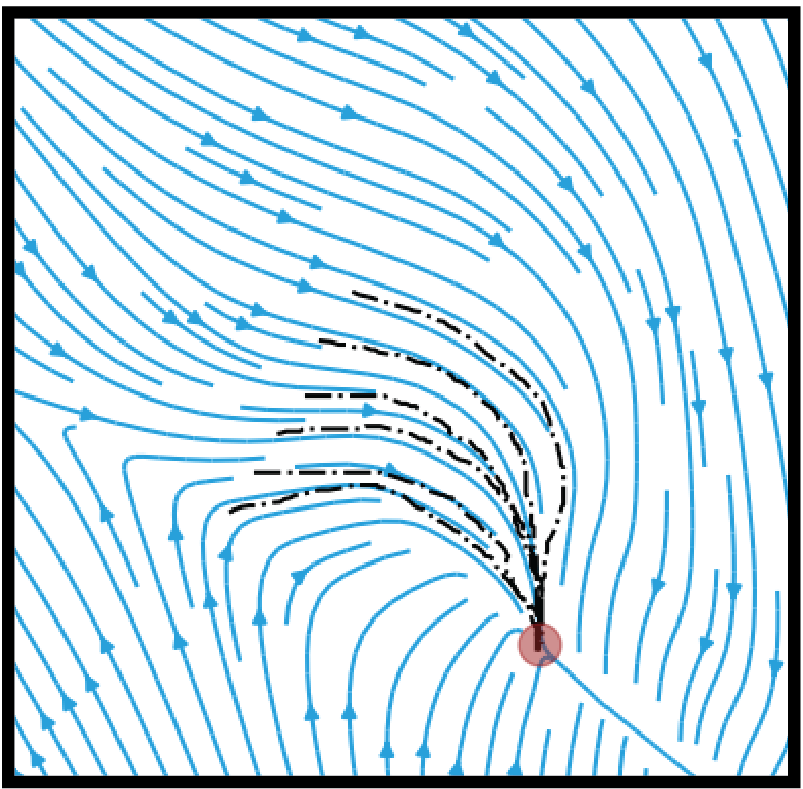}
		\caption{$\mathsf{C}_{\mathsf{pick}}, \mathsf{C}_{\mathsf{place}}$}
		\label{subFig:PlanarSkills}
	\end{subfigure}
	\begin{subfigure}[b]{0.13\textwidth}
		\includegraphics[width=\textwidth]{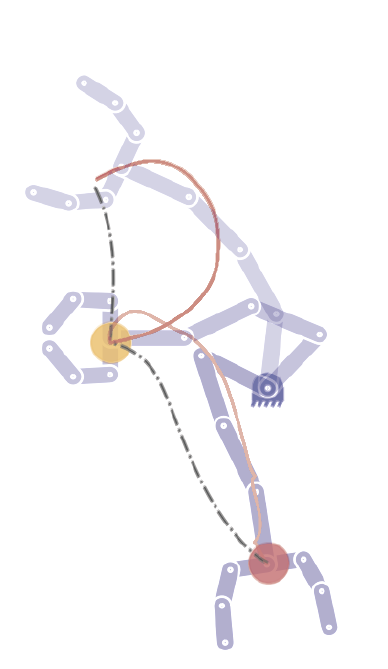}
		\caption{Diag. $\bm{W}(s)$}
		\label{subFig:PlanarDiagonal}
	\end{subfigure}
	\begin{subfigure}[b]{0.13\textwidth}
		\includegraphics[width=\textwidth]{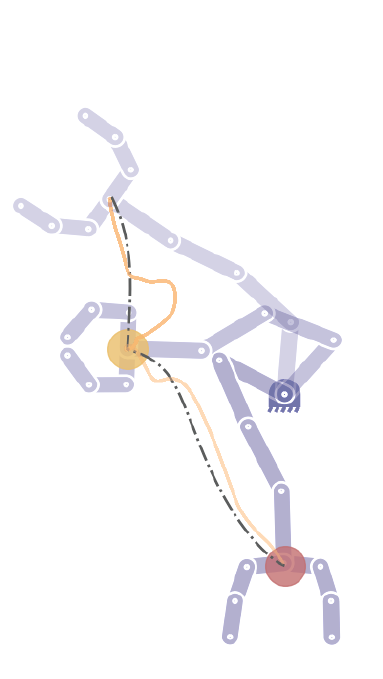}
		\caption{Full $\bm{W}(s)$}
		\label{subFig:PlanarFull}
	\end{subfigure}
	\begin{subfigure}[b]{0.13\textwidth}
		\includegraphics[width=\textwidth]{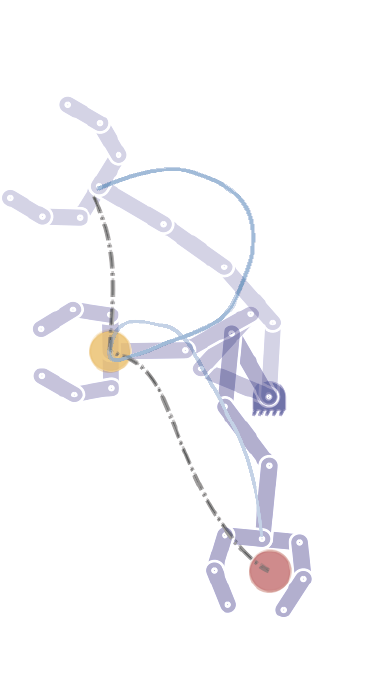}
		\caption{Baseline $\bm{W}$}
		\label{subFig:PlanarInterpolated}
	\end{subfigure}
	\begin{subfigure}[b]{0.13\textwidth}
		\includegraphics[width=\textwidth]{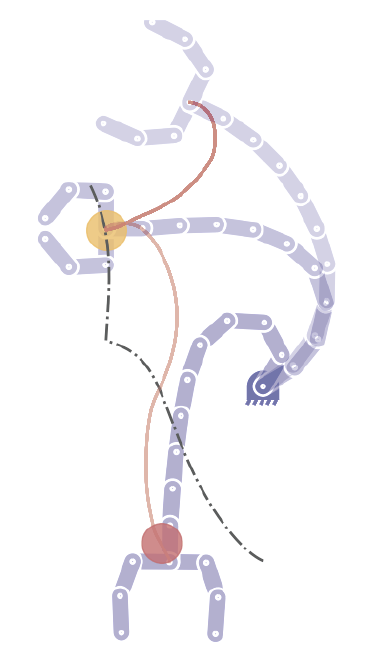}
		\caption{Generalization}
		\label{subFig:PlanarGeneralization}
	\end{subfigure}
	\begin{subfigure}[b]{0.29\textwidth}
		\includegraphics[width=.49\textwidth,trim={0 4.1cm 0 0},clip]{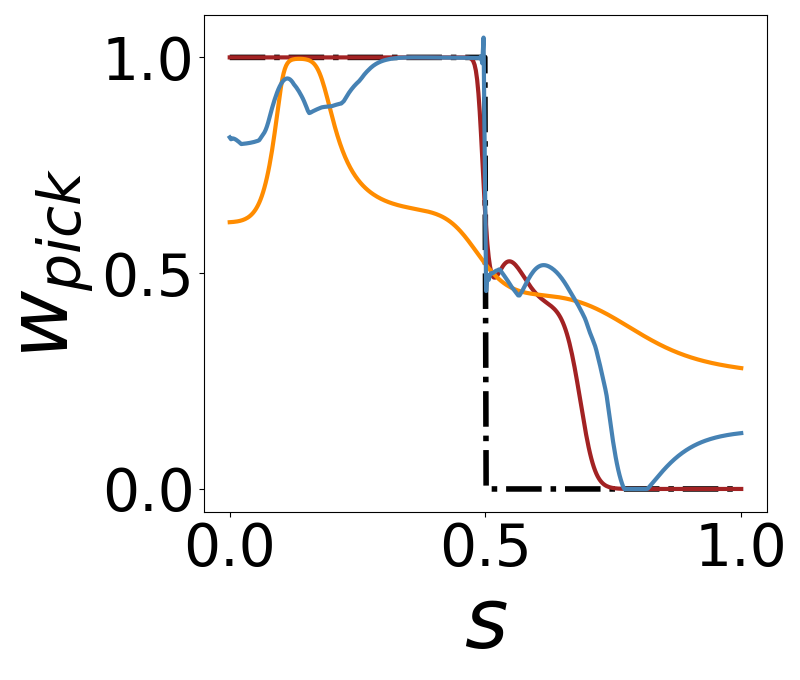}
		\includegraphics[width=.49\textwidth,trim={0 4.1cm 0 0},clip]{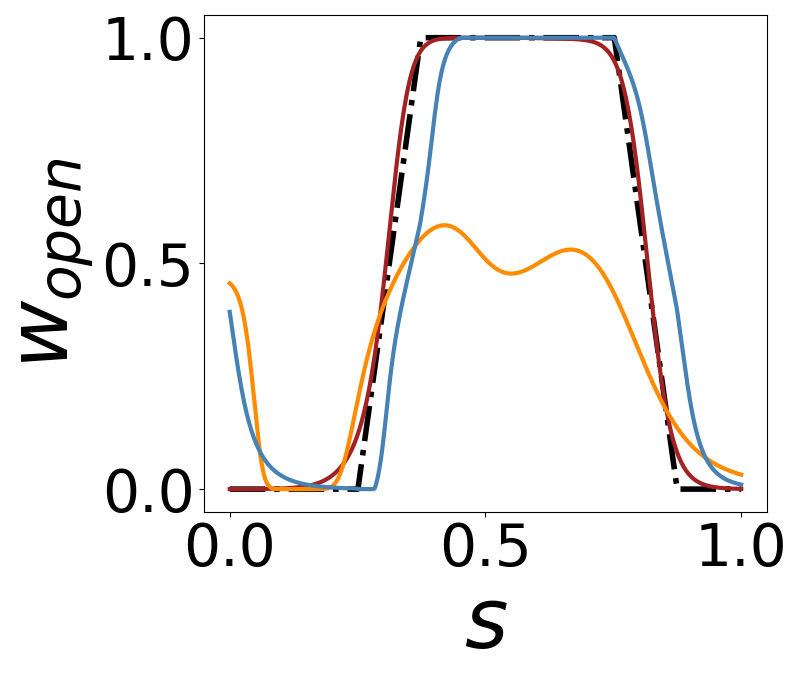}
		\includegraphics[width=.49\textwidth]{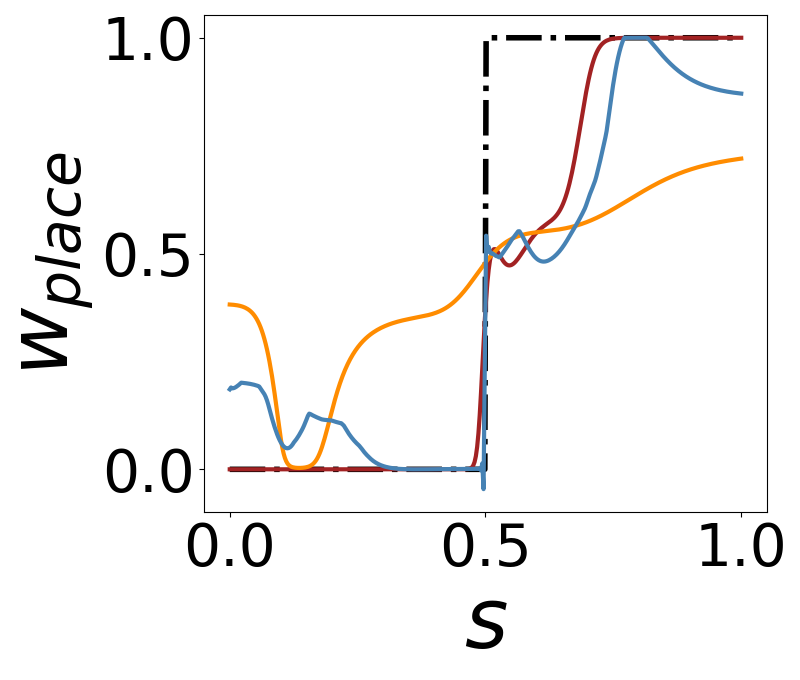}
		\includegraphics[width=.49\textwidth]{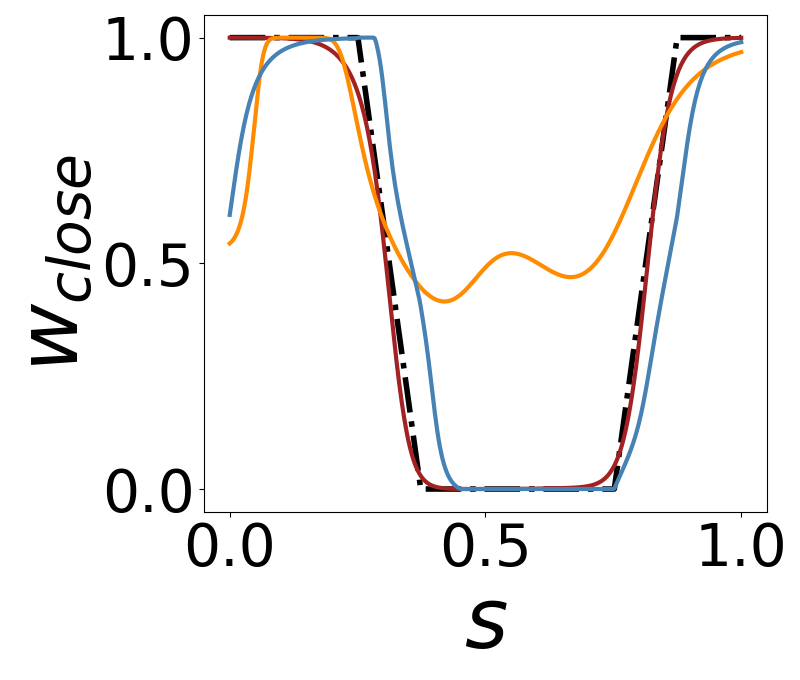}
		\caption{Evolution of $\text{diag}(\bm{W}(s))$}
		\label{subFig:PlanarWeights}
	\end{subfigure}
	\caption{Pick(\pickcircle)-and-place(\placecircle) task with planar robots. \emph{(a)} $\mathsf{Pick}$ (\emph{top}) and $\mathsf{place}$ (\emph{bottom}) DS skills (\blueDS). \emph{(b)}-\emph{(f)} Demonstration (\blackrepro), reproduction with the $4$-DoF robot using diagonal (\redrepro), full (\orangerepro) and baseline (\bluerepro) weights, and generalization with the $10$-DoF robot.}
	\label{Fig:PlanarExperiment}
	\vspace{-0.3cm}
\end{figure*}

In this section, we present our approach to sequence and blend manipulation skills. In the following, we assume a set of previously-defined individual robot skills  $\{\mathsf{C}_k\}_{k=1}^K$ (e.g., a skill library). The skills are considered as given black-box solutions, implying that their representations are unknown and may differ across the skills.
At each instant, each skill outputs a desired control value $\hat{\bm{\xi}}_k(\bm{\psi})$, depending on a current state $\bm{\psi}$, to be given to the robot in order to execute the skill. For example, dynamical-systems-based skills~\cite{Gribovskaya11:DS} provide a desired end-effector velocity depending on the current end-effector position, and time-dependent skills such as~\cite{Zhou2019:VMP} may output, e.g., a time-varying desired joint or end-effector position. The control values are specific to and may differ across skills.
We then consider a manipulation task consisting of an \emph{unknown} sequence of (some of) the aforementioned skills, possibly concurrently activated. We observe one or several \emph{optimal} demonstrations  $\{\tilde{\bm{\tau}}^{(d)}\}_{d=1}^D$ of the task consisting of the observed control values, i.e.,  $\tilde{\bm{\tau}}^{(d)} = [ \{\tilde{\bm{\xi}}_k^{(d)} (\bm{\psi}_s) \}_{k=1}^K ]_{s=0}^1$, where the phase variable $s\in [0, 1]$ encodes the task progress\footnote{In the remainder we drop dependencies on $\bm{\psi}_s$ to simplify the notation.}. In other words, $s=0$ and $s=1$ represent the beginning and the end of the task. 

\subsection{Illustrative example: pick-and-place with planar robots}
\label{subsec:IllustrativeExample}

For the sake of clarity of this section, the different concepts underlying our approach are introduced generally before being illustrated for a pick-and-place task executed by planar robots with grippers.
In this example, we observe a \emph{single} manually-designed demonstration $\tilde{\bm{\tau}}^{(1)}$ provided by a $4$-DoF teacher robot that picks an object, transports it, and places it at a given location (see  Fig.~\ref{Fig:PlanarExperiment}). The demonstration steps were achieved with proportional controllers activated using the weights of Fig.~\ref{subFig:PlanarWeights}.
We then consider a set of four skills $\{\mathsf{C}_{\mathsf{pick}}, \mathsf{C}_{\mathsf{place}}, \mathsf{C}_{\mathsf{open}}, \mathsf{C}_{\mathsf{close}}\}$, where the $\mathsf{pick}$/$\mathsf{place}$ and $\mathsf{open}$/$\mathsf{close}$ skills control the arm and gripper motion, respectively. 
Although we next disclose the skills types, remember that they are considered as given black-box solutions in our approach. Indeed, each skill only provides a desired control value $\hat{\bm{\xi}}_k(\bm{\psi}_s)$ depending on the state $\bm{\psi}_s$ at each task instant. 

The arm skills are encoded as dynamical systems (DS)~\cite{Gribovskaya11:DS} trained with the control Lyapunov function scheme of~\cite{Khansari14:DS}. 
The obtained DS, illustrated by Fig.~\ref{subFig:PlanarSkills}, can then be adapted to new situations via translations and rotations. The desired control values of the DS-based skills correspond to the end-effector velocity $\dot{\bm{p}}$ and depend on the current end-effector position $\bm{p}_s$, such that $\hat{\bm{\xi}}_{\mathsf{pick}}(\bm{\psi}_s) \equiv \hat{\dot{\bm{p}}}_{\mathsf{pick}}(\bm{p}_s)$ and $\hat{\bm{\xi}}_{\mathsf{place}}(\bm{\psi}_s) \equiv \hat{\dot{\bm{p}}}_{\mathsf{place}}(\bm{p}_s)$.
The desired control values of the gripper skills correspond to the velocity of the gripper joints $\dot{\bm{\gamma}}$. The velocities $\hat{\dot{\bm{\gamma}}}_{\mathsf{open}}$ and $\hat{\dot{\bm{\gamma}}}_{\mathsf{close}}$ are zero when the gripper is completely opened or closed, and constant otherwise, i.e., $\hat{\bm{\xi}}_{\mathsf{open}}(\bm{\psi}_s) \equiv \hat{\dot{\bm{\gamma}}}_{\mathsf{open}}(\bm{\gamma}_s)$ and $\hat{\bm{\xi}}_{\mathsf{close}}(\bm{\psi}_s) \equiv \hat{\dot{\bm{\gamma}}}_{\mathsf{close}}(\bm{\gamma}_s)$.
In this example, the phase variable $s$ is defined as $s=t/T$ with $t$ the elapsed time, and $T$ the total duration of the task. 

\subsection{Sequencing and blending of skills with QPs}
\label{subsec:QPformulation}
Similarly to multitask control, we propose to encode sequences of skills as QPs. 
Namely, given the desired control values $\{\hat{\bm{\xi}}_k\}_{k=1}^K$ output by the $K$ individual skills and the current control values $\{\bm{\xi}_k\}_{k=1}^K$, a sequence of skills can be generated by solving the following optimization problem
\begin{equation}
	\min_{\{\bm{\xi}_k\}_{k=1}^K} 
	\frac{1}{2}
	\left( \begin{smallmatrix}
	\hat{\bm{\xi}}_1 - \bm{\xi}_1 \\
	\vdots \\
	\hat{\bm{\xi}}_K - \bm{\xi}_K \\
	\end{smallmatrix} \right)^\trsp 
	\bm{W}(s) \left( \begin{smallmatrix}
	\hat{\bm{\xi}}_1 - \bm{\xi}_1 \\
	\vdots \\
	\hat{\bm{\xi}}_K - \bm{\xi}_K \\
	\end{smallmatrix} \right), 
	\label{Eq:SequenceQP}
\end{equation}
at each $s\in [0, 1]$, where $\bm{W}(s)$ is a varying weight matrix setting the relative importance of the skills throughout the sequence in function of the phase variable $s$ encoding the task progress. The problem~\eqref{Eq:SequenceQP} is usually augmented with linear constraints related to the robotic system (see \S~\ref{subsec:MultitaskControl}). In our case, we also include equality constraints for control values of the same type, i.e., $\bm{\xi}_i=\bm{\xi}_j$ if the skills $i$ and $j$ have the same type of outputs (e.g., both return end-effector pose values). For instance, following~\eqref{Eq:SequenceQP}, the optimization problem of our illustrative example is formulated as
\begin{align*}
\min_{\{ \dot{\bm{p}}_{\mathsf{pick}}, \dot{\bm{p}}_{\mathsf{place}} , \dot{\bm{\gamma}}_{\mathsf{open}}, \dot{\bm{\gamma}}_{\mathsf{close}}\}} 
&\frac{1}{2}
\left( \begin{smallmatrix}
\hat{\dot{\bm{p}}}_{\mathsf{pick}} - \dot{\bm{p}}_{\mathsf{pick}} \\
\hat{\dot{\bm{p}}}_{\mathsf{place}} - \dot{\bm{p}}_{\mathsf{place}}\\
\hat{\dot{\bm{\gamma}}}_{\mathsf{open}} - \dot{\bm{\gamma}}_{\mathsf{open}}\\
\hat{\dot{\bm{\gamma}}}_{\mathsf{close}} - \dot{\bm{\gamma}}_{\mathsf{close}} \\
\end{smallmatrix} \right)^\trsp 
\bm{W}(s) 
\left( \begin{smallmatrix}
\hat{\dot{\bm{p}}}_{\mathsf{pick}} - \dot{\bm{p}}_{\mathsf{pick}} \\
\hat{\dot{\bm{p}}}_{\mathsf{place}} - \dot{\bm{p}}_{\mathsf{place}}\\
\hat{\dot{\bm{\gamma}}}_{\mathsf{open}} - \dot{\bm{\gamma}}_{\mathsf{open}}\\
\hat{\dot{\bm{\gamma}}}_{\mathsf{close}} - \dot{\bm{\gamma}}_{\mathsf{close}} \\
\end{smallmatrix} \right), \\ &\text{ s.t. }\; \dot{\bm{p}}_{\mathsf{pick}}=\dot{\bm{p}}_{\mathsf{place}} \; \text{ and }\; \dot{\bm{\gamma}}_{\mathsf{open}} = \dot{\bm{\gamma}}_{\mathsf{close}}.
\end{align*}
The constraints come from the shared control values across skills, i.e., the end-effector velocity $\dot{\bm{p}}$, and the gripper joints velocity $\dot{\bm{\gamma}}$ for the $\mathsf{pick}/\mathsf{place}$ and $\mathsf{open}/\mathsf{close}$ skills, respectively. These constraints can directly be integrated into the optimization problem, which is equivalently written as
\begin{equation}
\min_{\{\dot{\bm{p}}, \dot{\bm{\gamma}}\}} 
\frac{1}{2}
\left( \begin{smallmatrix}
\hat{\dot{\bm{p}}}_{\mathsf{pick}} - \dot{\bm{p}} \\
\hat{\dot{\bm{p}}}_{\mathsf{place}} - \dot{\bm{p}}\\
\hat{\dot{\bm{\gamma}}}_{\mathsf{open}} - \dot{\bm{\gamma}}\\
\hat{\dot{\bm{\gamma}}}_{\mathsf{close}} - \dot{\bm{\gamma}} \\
\end{smallmatrix} \right)^\trsp 
\bm{W}(s) 
\left( \begin{smallmatrix}
\hat{\dot{\bm{p}}}_{\mathsf{pick}} - \dot{\bm{p}} \\
\hat{\dot{\bm{p}}}_{\mathsf{place}} - \dot{\bm{p}}\\
\hat{\dot{\bm{\gamma}}}_{\mathsf{open}} - \dot{\bm{\gamma}}\\
\hat{\dot{\bm{\gamma}}}_{\mathsf{close}} - \dot{\bm{\gamma}} \\
\end{smallmatrix} \right).
\label{Eq:IllustrativeQPintegratedconstraints}
\end{equation}

Note that~\eqref{Eq:SequenceQP} can be equivalently formulated as~\eqref{Eq:QP} with the optimization variable $\bm{z}= \left(\begin{smallmatrix}
\bm{\xi}_1^\trsp & \ldots &  \bm{\xi}_K^\trsp
\end{smallmatrix}\right)^\trsp$, and cost parameters $\bm{Q} = \bm{W}$, $\bm{c}=-\bm{W}\hat{\bm{z}}$ with $\hat{\bm{z}}= \left(\begin{smallmatrix}
\hat{\bm{\xi}}_1^\trsp & \ldots &  \hat{\bm{\xi}}_K^\trsp
\end{smallmatrix}\right)^\trsp$. Importantly, the skill ordering in~\eqref{Eq:SequenceQP} is arbitrary.
Indeed, the sequence is defined by the weight matrix, that is learned from demonstrations, as explained next. 
Skills can be added by extending $\hat{\bm{z}}$ with their control values and expanding $\bm{W}$ accordingly.

\begin{figure}
	\centering
	\includegraphics[width=0.5\textwidth]{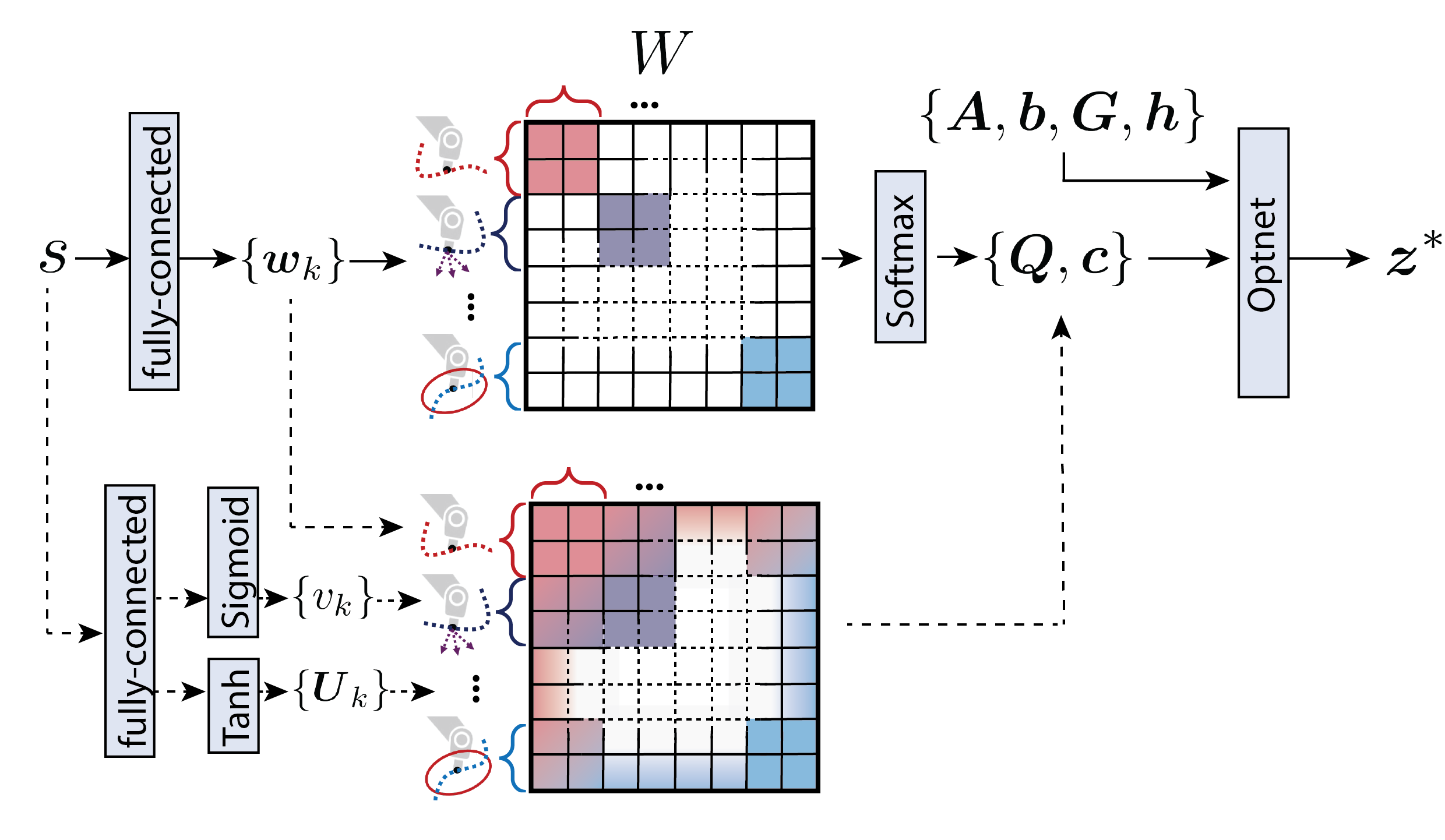}
	\caption{Illustration of the proposed learning approach. The relative importance of the skills is encoded by $\bm{W}$ as a function of $s$. An Optnet layer, solving a QP whose parameters depend on $\bm{W}$, is then used to determine the control command $\bm{z}^*$.
		$\bm{W}$ is either a block-diagonal (\emph{top}), or a full (\emph{bottom}) matrix. The dashed arrows are only activated in the latter to learn the off-diagonal elements.}
	\label{Fig:SequencingBlendingNetwork}
	\vspace{-0.2cm}
\end{figure}

Given one or several $D$ demonstrations $\{\tilde{\bm{\tau}}^{(d)}\}_{d=1}^D$ of a manipulation task, we aim at learning the skill weight function $s \mapsto \bm{W}(s): \mathbb{R} \to \mathcal{S}_{\ty{+}}^n$, so that the reproduction $\bm{\tau}=[\{\bm{\xi}_{k,s}^*\}_{k=1}^K ]_{s=0}^1$, i.e., the sequence of skills obtained by solving~\eqref{Eq:SequenceQP} for $s\in[0, 1]$, replicates the demonstrated task. This corresponds to minimizing a loss function $\ell(\bm{\tau}, \{\tilde{\bm{\tau}}^{(d)}\}_{d=1}^D)$ measuring the quality of the reproduction.
To do so, we need to solve a nested optimization: For each time instance of the task, we solve~\eqref{Eq:SequenceQP}, and the whole set of solutions $[\{\bm{\xi}_{k,s}^*\}_{k=1}^K ]_{s=0}^1$ is then used to minimize the loss $\ell(\bm{\tau}, \{\tilde{\bm{\tau}}^{(d)}\}_{d=1}^D)$. To solve this problem, we leverage Optnet~\cite{Amos17:Optnet} to integrate the QP~\eqref{Eq:SequenceQP} into a neural network. Optnet allows us (\emph{i}) to represent the QP parameters as functions, and (\emph{ii}) to differentiate $\ell$ with respect to the QP parameters to solve the outer optimization of our nested problem using gradient-based approaches. In other words, Optnet backpropagates the loss $\ell$ to optimize both the phase-dependent skills weights $\bm{W}(s)$ and the control outputs $\bm{z}$.
Thus, we can learn the relative importance of the skills throughout the task execution via the matrix $\bm{W}(s)$. 
Our proposed neural network takes the phase variable $s$ as input, and consists of (\emph{i}) a fully-connected layer coupled with a softmax activation function, whose outputs are the QP parameters $\{\bm{Q},\bm{c},\bm{A},\bm{b},\bm{G}, \bm{h}\}(s)$ (see \S~\ref{subsec:SPDweights} for details), 
and (\emph{ii}) of an Optnet layer~\eqref{Eq:Optnet}, where $\bm{z}_i=s$, and $\bm{z}_{i+1}=\bm{z}*$ is the control command transmitted to the robot to execute the task. Our approach is illustrated by Fig.~\ref{Fig:SequencingBlendingNetwork}.

It is important to emphasize that the proposed formulation not only learns sequences of skills, but also blends the transition between individual skills "for free". Indeed, the coupling of the fully-connected layer with a softmax activation induces smooth non-binary weight functions $\bm{W}(s)$, therefore leading to smooth variations of the relative importance of the skills, i.e., to smooth transitions. This allows our neural architecture to learn and reproduce seamless transitions, as usually observed in human demonstrations.
This also implies that skills are not necessarily executed in a strict sequence, but may be activated concurrently if required by the task.

The individual skills outputs $\hat{\bm{\xi}}_k$ may be defined either in task space (e.g., end-effector pose, or velocity), or in joint space (e.g., joint position, or velocity). In the former case, it may be desirable to directly solve the optimization~\eqref{Eq:SequenceQP} with respect to joint variables when executing the reproduction on the robot. To do so, the current control values $\{\bm{\xi}_k\}_{k=1}^K$ can be expressed in function of the joint values by exploiting the kinematic or dynamic relationship between the task- and joint-space variables. 
In our illustrative example, this corresponds to solving, during the reproduction,
\begin{equation}
\min_{\{\dot{\bm{\alpha}}, \dot{\bm{\gamma}}\}} 
\frac{1}{2}
\left( \begin{smallmatrix}
\hat{\dot{\bm{p}}}_{\mathsf{pick}} - \bm{J}\dot{\bm{\alpha}} \\
\hat{\dot{\bm{p}}}_{\mathsf{place}} - \bm{J}\dot{\bm{\alpha}}\\
\hat{\dot{\bm{\gamma}}}_{\mathsf{open}} - \dot{\bm{\gamma}}\\
\hat{\dot{\bm{\gamma}}}_{\mathsf{close}} - \dot{\bm{\gamma}} \\
\end{smallmatrix} \right)^\trsp 
\bm{W}(s) 
\left( \begin{smallmatrix}
\hat{\dot{\bm{p}}}_{\mathsf{pick}} - \bm{J}\dot{\bm{\alpha}} \\
\hat{\dot{\bm{p}}}_{\mathsf{place}} - \bm{J}\dot{\bm{\alpha}}\\
\hat{\dot{\bm{\gamma}}}_{\mathsf{open}} - \dot{\bm{\gamma}}\\
\hat{\dot{\bm{\gamma}}}_{\mathsf{close}} - \dot{\bm{\gamma}} \\
\end{smallmatrix} \right),
\label{Eq:IllustrativeQPrepro}
\end{equation}
where the arm skills outputs are expressed as $\dot{\bm{p}} = \bm{J}\dot{\bm{\alpha}}$ with $\dot{\bm{\alpha}}$ and $\dot{\bm{\gamma}}$ the arm and gripper joint velocities, respectively, and $\bm{J}$ the manipulator Jacobian. 
Finally, note that nonlinear relationships must be linearized for the QP formulation.

\subsection{Definition of the loss function}
\label{subsec:Loss}
In this section, we take inspiration from the IOC approach of~\cite{Englert17:InverseKKT} to define the loss function $\ell$ used to train the neural network previously introduced. Namely, we assume that the demonstrations $\{\tilde{\bm{\tau}}^{(d)}\}_{d=1}^D$ are optimal, i.e., they are optimal solutions to the QP problem~\eqref{Eq:SequenceQP} and thus satisfy its KKT conditions. As the QP constraints are satisfied during optimal demonstrations, the KKT conditions (\emph{ii})-(\emph{v}) are automatically fulfilled. Therefore, determining the optimal parameters $\bm{\theta}^*$ of our neural network can be understood as searching for the parameters $\bm{\theta}$ fulfilling the first KKT condition for all the demonstrations. This corresponds to minimizing the loss 
\begin{align}
\ell(\bm{\tau} (\bm{\theta}), \{\tilde{\bm{\tau}}^{(d)}\}_{d=1}^D) &= \sum_{d=1}^{D} \ell^{(d)}(\bm{\theta}) 
\;\;  \label{Eq:Loss} \\ \text{ with }\;\; 
\ell^{(d)}(\bm{\theta}) &= \sum_{s} \| \nabla_{\bm{z}} \mathcal{L}(s, \bm{\theta}, \bm{z}, \tilde{\bm{z}}^{(d)}, \bm{\lambda}^{(d)})\|^2, \nonumber
\end{align}
where we sum over the demonstrations and the progress of the task via the phase variable $s$. The Lagrangian of the problem~\eqref{Eq:SequenceQP} and its derivative for the $d$-th demonstration are 
\begin{align*}
\mathcal{L}(s, \bm{\theta}, \bm{z}, \tilde{\bm{z}}^{(d)}, \bm{\lambda}^{(d)}) = &
\frac{1}{2}
\big( \tilde{\bm{z}}^{(d)}_s - \bm{z}_s \big)^\trsp 
\bm{W}_s(\bm{\theta}) \big( \tilde{\bm{z}}^{(d)}_s - \bm{z}_s \big) \nonumber \\ &+ \bm{\lambda}^{(d)\trsp}_s \left(\bm{P}_s\bm{z}_s - \bm{r}_s\right), 
\end{align*}
\begin{equation*}
\nabla_{\bm{z}} \mathcal{L}(s, \bm{z}, \bm{\theta}, \tilde{\bm{z}}^{(d)}, \bm{\lambda}^{(d)}) =  
\bm{W}_s(\bm{\theta}) \big( \tilde{\bm{z}}_s^{(d)} - \bm{z}_s \big) + \bm{P}_s^\trsp \bm{\lambda}_s^{(d)},
\end{equation*}
where $\bm{z}_s \equiv \bm{z}(s)$, $\tilde{\bm{z}}^{(d)}= \left(\begin{smallmatrix}
\tilde{\bm{\xi}}_1^{(d)\trsp} & \ldots &  \tilde{\bm{\xi}}_K^{(d)\trsp}
\end{smallmatrix}\right)^\trsp$ is the vector of demonstrated skills outputs, $\bm{P} = \left( \begin{smallmatrix}
\bm{A} \\ \bm{G} \end{smallmatrix} \right)$ and $\bm{r} = \left( \begin{smallmatrix}
\bm{b} \\ \bm{h} \end{smallmatrix} \right)$ are the stacked constraints parameters, and $\bm{\lambda} = \left( \begin{smallmatrix}
\bm{\mu} \\ \bm{\nu} \end{smallmatrix} \right)$ is the vector of Lagrangian multipliers. Moreover, we can express $\bm{\lambda}^{(d)}$ in function of $\bm{\theta}$ for each demonstration $d$ by minimizing the loss $\ell^{(d)}$ subject to the KKT complementary condition, i.e., $\nabla_{\bm{\lambda}^{(d)}}\ell^{(d)}(\bm{\theta}, \bm{\lambda}^{(d)})=0$. Therefore, by setting the optimization variable $\bm{z}_s$ to the output $\bm{z}_s^*(\bm{\theta})$ of our network, the loss of each demonstration is \footnote{Equivalently, $\ell^{(d)}(\bm{\theta}) \!=\!  \sum_{s} \| \bm{W}_s(\bm{\theta}) ( \tilde{\bm{z}}_s^{(d)} - \bm{z}_s^*(\bm{\theta}) ) \|^2$ for constant $\bm{P}_s$.}    
\small
\begin{equation*}
\ell^{(d)}(\bm{\theta}) =  \sum_{s}
\|
\Big( \bm{I} - \bm{P}_s^\trsp \big( \bm{P}_s\bm{P}_s^\trsp \big)^{-1} \bm{P}_s\Big)\bm{W}_s(\bm{\theta}) \big( \tilde{\bm{z}}_s^{(d)} - \bm{z}_s^*(\bm{\theta}) \big) \|^2. 
\end{equation*}
\normalsize
The loss~\eqref{Eq:Loss} inherently includes the task specifications via the demonstrations and the QP KKT conditions, and does not require additional task-specific design.
To avoid the singular solution $\bm{W}_s(\bm{\theta}) = \bm{0} \; \forall s$, we leverage the softmax activation function, as explained next. Thus, at least one skill is given a high relative importance at each instant of the task.

\subsection{Skills weights as positive-semidefinite matrices}
\label{subsec:SPDweights} 
As mentioned previously, the QP parameters are determined by the first part of our neural network. Specifically, the cost parameters are $\bm{Q}_s(\bm{\theta}) = \bm{W}_s(\bm{\theta})$, $\bm{c}_s(\bm{\theta})=-\bm{W}_s(\bm{\theta})\hat{\bm{z}}_s$ where the weight matrix $\bm{W}_s(\bm{\theta})$ is learned by the network. The constraints parameters relate to skills outputs and to the robot physical characteristics. To obtain valid QPs, or equivalently to prevent skills to have negative relative importance weights, the weight matrices must be PSD, i.e., $\bm{W}\in\mathcal{S}_{\ty{+}}^n$. We here describe two approaches to learn PSD weight matrices. 

\paragraph{Diagonal weight matrices} In this case, we define
\begin{equation}
\bm{W}(\bm{\theta}) = \text{diag} \left(\begin{matrix}
w_{1}(\bm{\theta}) \bm{I}_1, & \ldots&,  w_K(\bm{\theta}) \bm{I}_K
\end{matrix}\right), 
\label{Eq:DiagWeightMatrix}
\end{equation}
where each block $w_k(\bm{\theta}) \bm{I}_k$ weights the output of the $k$-th skill, and the scalars $\{w_k(\bm{\theta})\}_{k=1}^K$ are obtained from the fully-connected layer followed by a softmax activation function. The latter ensures that the scalar weights are positive and sum to $1$, thus guaranteeing that $\bm{W}$ is PSD, and that at least one skill is activated at any instant of the task. Notice that we defined the different blocks as proportional to identity matrices to avoid altering the outputs of individual skills.

\paragraph{Full weight matrices} Such matrices allow us to express correlations between different skills, i.e, between their control values $\hat{\bm{\xi}}$, throughout the task. This naturally occurs in various tasks. For example, when approaching and grasping an object, the hand closure is correlated with the velocity at which the object is approached. We learn matrices
\begin{equation}
	\bm{W}(\bm{\theta}) = \left( \begin{smallmatrix} 
	w_1(\bm{\theta})\bm{I}_1 & \bm{W}_{12}(\bm{\theta}) & \ldots  & \bm{W}_{1K}(\bm{\theta}) \\ 
	\bm{W}_{12}^\trsp(\bm{\theta}) &  w_2(\bm{\theta})\bm{I}_2 & \ldots &  \bm{W}_{2K}(\bm{\theta})\\ 
	\vdots & \vdots & \ddots & \vdots \\ \bm{W}_{1K}^\trsp(\bm{\theta}) & \bm{W}_{2K}^\trsp(\bm{\theta}) & \ldots & w_K(\bm{\theta})\bm{I}_K
	\end{smallmatrix} \right),
\label{Eq:FullWeightMatrix}
\end{equation}
where the off-diagonal blocks $\bm{W}_{jk}$ encode the correlations between the outputs of the skills $j$ and $k$. To guarantee the positive semidefiniteness of the matrices $\bm{W}$, we propose to learn the diagonal and off-diagonal blocks separately. Firstly, the scalar terms $\{w_k(\bm{\theta})\}_{k=1}^K$ are obtained as described in the previous paragraph. Secondly, the off-diagonal matrices $\{\bm{W}_{jk}(\bm{\theta})\}_{j,k=1}^K$ are obtained by leveraging the properties of matrices with positive block-diagonal elements~\cite{Bhatia07:PDmatrices}, namely
\begin{equation}
	\left( \begin{smallmatrix}
	\bm{Y} & \bm{X} \\ \bm{X}^\trsp & \bm{Z}
	\end{smallmatrix} \right) \in\mathcal{S}_{\ty{+}}^n \iff \bm{X}=\bm{Y}^{1/2} \bm{K} \bm{Z}^{1/2}, 
	\label{Eq:PropertyBlockSPD}
\end{equation}
where $\bm{K}$ is a contraction matrix, i.e., $\|\bm{K}\| \leq 1$.
Therefore, we use a second fully-connected layer to learn the contraction matrices as $\bm{K}_k = v_k(\bm{\theta}) \frac{\bm{U}_k(\bm{\theta})}{\|\bm{U}_k(\bm{\theta})\|}$, with a tanh and a sigmoid activation function applied to $\bm{U}_k$ and $v_k$, respectively, so that $v_k\in [0,1]$. The off-diagonal elements $\{\bm{W}_{jk}(\bm{\theta})\}_{j,k=1}^K$ are then computed recursively using the right-hand side of~\eqref{Eq:PropertyBlockSPD}. For instance, in the case of a matrix composed of 3 skills, we first compute $\bm{X}=\bm{W}_{12}$ with $\bm{Y}=w_1\bm{I}_1$ and $\bm{Z} = w_2\bm{I}_2$, and then $\bm{X}=\left(\begin{smallmatrix}
\bm{W}_{13}^\trsp & \bm{W}_{23}^\trsp
\end{smallmatrix} \right)^\trsp$ with $\bm{Y}= \left( \begin{smallmatrix} 
w_1\bm{I}_1 & \bm{W}_{12} \\ 
\bm{W}_{12}^\trsp &  w_2\bm{I}_2 \end{smallmatrix} \right)
$ and $\bm{Z} = w_3\bm{I}_3$.
Note that, to facilitate the training of full weight matrices, we initialize the parameters $\bm{\theta}$ of the scalar terms $\{w_k(\bm{\theta})\}_{k=1}^K$ with a previously-trained diagonal model.


\section{Experiments}
\label{sec:Experiments}
In this section, we evaluate our approach with different robotic platforms and manipulation tasks. 
All computations were performed on a laptop with $2.60$GHz $\times 12$ CPU and $31$ GiB RAM.
A video of the experiments accompanies the paper (\url{https://youtu.be/00NXvTpL-YU}\normalsize), and source codes are available at \url{https://github.com/NoemieJaquier/sequencing-blending/}.

\subsection{Illustrative example: pick-and-place with planar robots}
\label{subsec:PlanarExample}
We first consider the pick-and-place task introduced in \S~\ref{subsec:IllustrativeExample} and train our approach using diagonal and full weight matrices on the provided single manually-designed demonstration. In order to guarantee that one arm and one hand skill are activated at each instant of the task, we use one softmax activation function for each of the arm and gripper pairs of skills, namely $\mathsf{pick}/\mathsf{place}$ and $\mathsf{open}/\mathsf{close}$.
The task is then reproduced by the 4-DoF robot. As a baseline, we consider the case where the QP~\eqref{Eq:IllustrativeQPintegratedconstraints} with diagonal weights does not require additional constraints, so that its solution is $\dot{\bm{p}}^*=w_{\mathsf{pick}}\hat{\dot{\bm{p}}}_{\mathsf{pick}} + w_{\mathsf{place}}\hat{\dot{\bm{p}}}_{\mathsf{place}}$, $\dot{\bm{\gamma}}^* = w_{\mathsf{open}} \hat{\dot{\bm{\gamma}}}_{\mathsf{open}} + w_{\mathsf{close}} \hat{\dot{\bm{\gamma}}}_{\mathsf{close}}$. In this case, as the QP solution is readily available, we do not need to solve a nested optimization to minimize a given loss. Instead, the loss~\eqref{Eq:Loss} can be minimized independently for each value of $s$ with classical optimization methods. Finally, a $10$-DoF student robot is requested to reproduce the learned sequence of skills with different pick and place positions. To do so, the $\mathsf{pick}$ and $\mathsf{place}$ DS skills are adapted to the new target points. For all reproductions, the QP is solved with respect to the arm and gripper joint velocities using~\eqref{Eq:IllustrativeQPrepro}. 

Fig.~\ref{subFig:PlanarDiagonal} depicts the demonstrated trajectory, as well as the reproduction of the task by the $4$-DoF robot with a diagonal weight matrix. Our approach successfully sequences the available skills and reproduces the task by picking and placing the object at the required locations. The differences of trajectory between the demonstration and the reproduction are due to the fact that the DS arm skills naturally follow a different trajectory than the demonstration between the target points (remember that the demonstration was generated independently from the given skills). For the same reason, the learned weights slightly differ from the manually-designed ones used to generate the demonstration (Fig.~\ref{subFig:PlanarWeights}). The differences of trajectory are attenuated when using a full weight matrix (see Fig.~\ref{subFig:PlanarFull}), where correlations between skills are exploited to better match the demonstration. Note that only the diagonal weights are represented in Fig.~\ref{subFig:PlanarWeights}. 
As expected, the baseline looks similar to our approach with diagonal weight matrix (see Fig.~\ref{subFig:PlanarInterpolated}). Slight differences may be due to the different optimizations and to local minima in the loss. However, notice that the baseline applies only to very simple QPs, which are unrealistic for most applications (incl. for the experiments of \S~\ref{subsec:PouringTask}-~\ref{subsec:MMMexperiment}). Also, in contrast to our approach, the baseline does not learn the weight matrix as a parametric function of the phase variable.  
Fig.~\ref{subFig:PlanarGeneralization} depicts the reproduction of the learned sequence by the $10$-DoF robot using a diagonal weight matrix, showing that our approach successfully generalizes to different pick and place locations. As the full weight matrix naturally overfits a single demonstration, it is not well suited to generalize in this case. 

\subsection{Pouring task with a humanoid robot}
\label{subsec:PouringTask}
\begin{figure*}
	\centering
	\includegraphics[width=.8\textwidth]{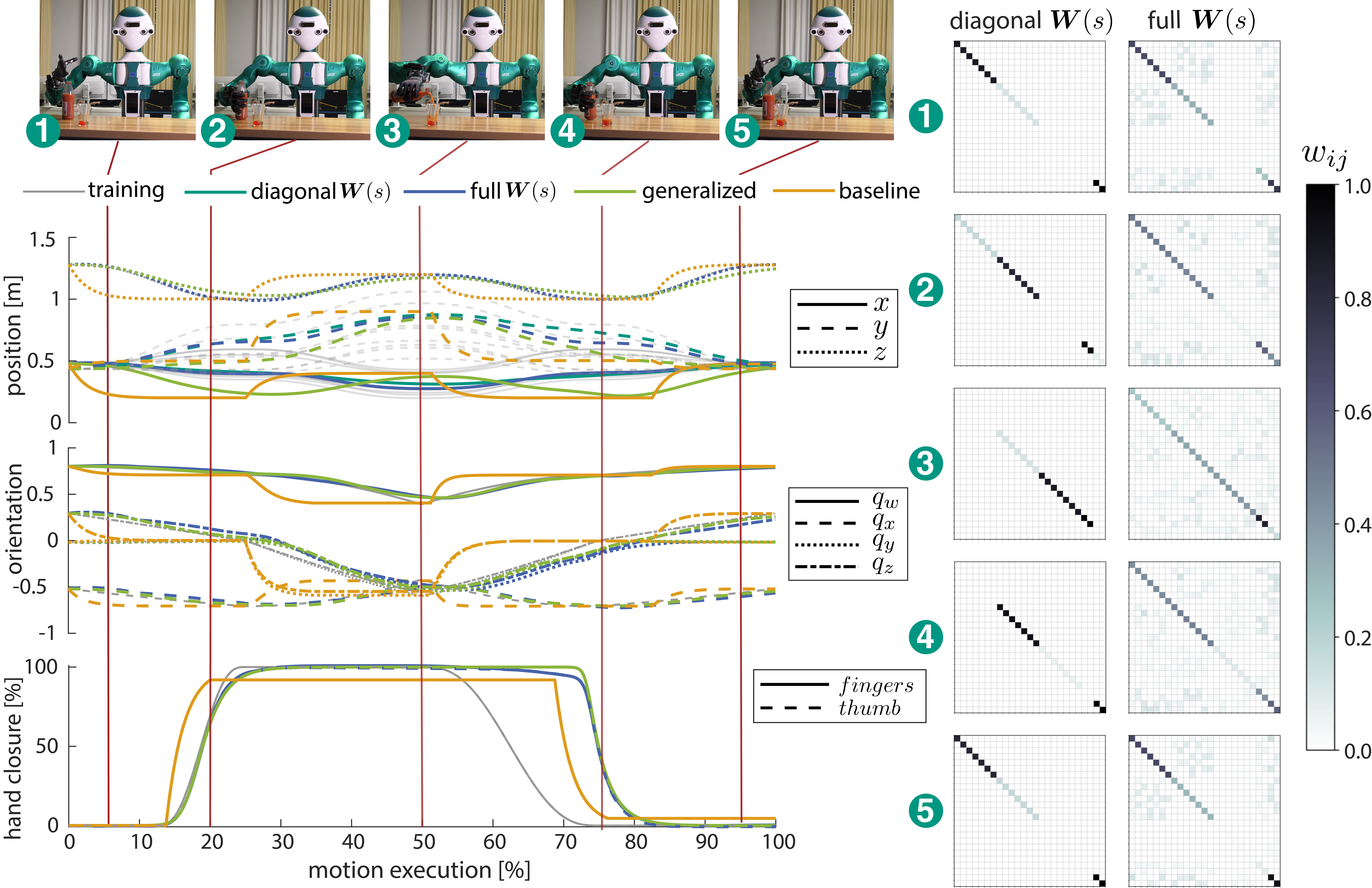} 
	\caption{Pouring task with a humanoid robot. The \emph{top} row shows snapshots of the robot in the resting position (\emph{1}) and executing the $\mathsf{approach}$ (\emph{2}), $\mathsf{pour}$ (\emph{3}), $\mathsf{place}$ (\emph{4}) and $\mathsf{retreat}$ (\emph{5}) skills during the task. The \emph{bottom-left} graphs depict the demonstrated (\graytraining) and reproduced hand position, orientation, and closure trajectories. Reproductions are obtained with our approach using diagonal (\kitgreenrepro) and full  (\kitbluerepro) weight matrices. A generalized motion obtained with diagonal weight matrices (\kitlightgreenrepro), as well as a baseline where skills are manually sequenced without blending (\kitorangerepro), are also displayed. The \emph{right} column depicts the learned diagonal and full weight matrices at different task instants.}
	\label{Fig:RobotExperiment}
	\vspace{-0.2cm}
\end{figure*}
Here, we apply our approach in a real-world scenario to learn a complex sequence of skills on the humanoid robot ARMAR-6~\cite{Asfour19:Armar6}. The robot is positioned in front of a table, on which are placed an empty glass and a $1$-liter plastic bottle partially filled with orange juice. 
The scenario consists of a pouring task, where the robot grasps the bottle, pours juice into the glass, and places the bottle back on the table. 
The positions of the objects are assumed a priori known by the robot, but could equally be inferred by a perception system.

As for the previous experiment, a set of skills is provided as black-box solutions. Specifically, four skills are defined for the arm, namely $\mathsf{approach}$ the bottle, $\mathsf{pour}$, $\mathsf{place}$ the bottle back, and $\mathsf{retreat}$ the arm. Moreover, two joint-velocity-based skills are provided for the five-fingered hand, namely $\mathsf{open}$ and $\mathsf{close}$ in a power cylindrical grasp. The four arm skills $\{\mathsf{C}_{\mathsf{approach}},\mathsf{C}_{\mathsf{pour}}, \mathsf{C}_{\mathsf{place}},\mathsf{C}_{\mathsf{retreat}}\}$ are defined by DS with radial vector fields pointing toward a fixed point attractor. Their desired control values correspond to the end-effector linear and angular velocities $\hat{\dot{\bm{p}}}$ and $\hat{\dot{\bm{q}}}$, which depend on the current end-effector position $\bm{p}_s\in\mathbb{R}^3$ and orientation $\bm{q}_s\in\mathcal{S}^3$,
i.e., $\hat{\bm{\xi}}(\bm{\psi}_s) \equiv \left(\begin{smallmatrix}
\hat{\dot{\bm{p}}}(\bm{p}_s) \\ \hat{\dot{\bm{q}}}(\bm{q}_s)
\end{smallmatrix}\right)$.
The fixed point attractors of the four arm skills are the robot hand grasp pose on the bottle for the $\mathsf{approach}$ skill, a tilted hand pose above the glass for the $\mathsf{pour}$ skill, the hand pose at the position of the bottle on the table for the $\mathsf{place}$ skill, and the hand resting pose for the $\mathsf{retreat}$ skill.
The hand skills $\{\mathsf{C}_{\mathsf{open}},\mathsf{C}_{\mathsf{close}}\}$ are defined similar to the gripper skills of the pick-and-place example, and thus open and close all finger joints by controlling their velocity.
We train our approach on seven manually-designed demonstrations for which an operator defined the arm and hand trajectories. The bottle and glass positions were varied of $\pm 10$ and $\pm 20$ cm along the $x$ and $y$ axes, respectively. 
As previously, we use two softmax activation functions for the arm and hand skills, and the phase variable is $s=t/T$.

After the learning phase, the robot successfully reproduced the pouring task using both diagonal and full weight matrices (see Fig.~\ref{Fig:RobotExperiment} (\emph{top-left})). Moreover, our approach not only succeeded at learning the desired sequence of skills, but also resulted in seamless transitions as indicated by the absence of pauses and by the smoothness of the trajectories depicted in~Fig.~\ref{Fig:RobotExperiment} (\emph{bottom-left}). 
The learned weight matrices are represented in Fig.~\ref{Fig:RobotExperiment} (\emph{right}) for the diagonal and full cases. Although the resulting trajectories look similar, the matrices still differ in the relative importance attributed to each skill. Notably, the model with full weight matrices exploits the correlation between the skills to shape the reproduced trajectory, thus featuring lower diagonal values than the diagonal model. Therefore, full weight matrices have better representation capabilities than their diagonal counterpart. However, this comes at the expense of generalization abilities. 
Indeed, as shown in~Fig.~\ref{Fig:RobotExperiment}, the diagonal model was able to generalize to bottle and glass locations that were outside the demonstrated range (here, the bottle and glass positions were swapped along the $x$ axis), which the full model could only achieve for locations close to the demonstrations. 
Finally, we compared our approach to a baseline obtained by manually sequencing the given skills without any learning or blending. As shown in~Fig.~\ref{Fig:RobotExperiment} (\emph{bottom-left}), the baseline trajectory is characterized by obvious jerky transitions. The resulting timing would cause the robot to overfill the glass, thus failing the reproduction.
Importantly, our approach is well-suited for learning and executing the sequence of skills on a real robot. Indeed, the pouring task training lasted a couple of minutes, and the testing time was $\sim3$-$4$ ms per timestamp, which allowed us to execute our approach at a control frequency of $200$ Hz.

\subsection{Bimanual sweeping task learned from human data}  
\label{subsec:MMMexperiment}
\begin{figure*}
	\centering
	\begin{subfigure}[b]{0.19\textwidth}
		\centering
		\includegraphics[width=.9\textwidth, trim={0 0.25cm 0 0.25cm}]{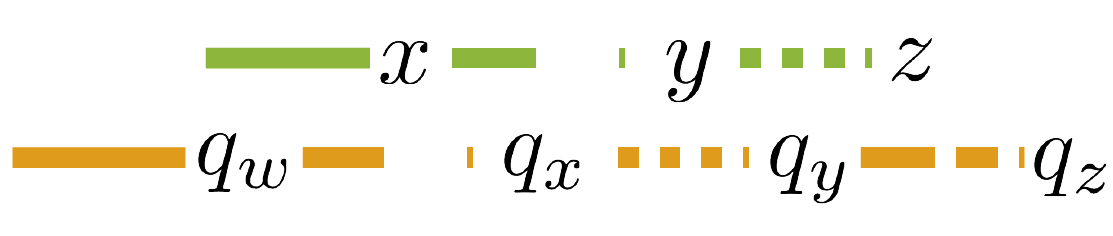} 
		\includegraphics[width=.08\textwidth, trim={0.1cm 2cm 0.1cm 2cm}]{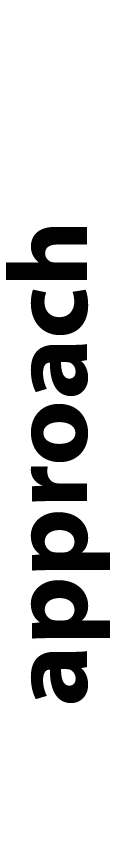}
		\includegraphics[width=.88\textwidth, trim={0 1.6cm 0 0},clip]{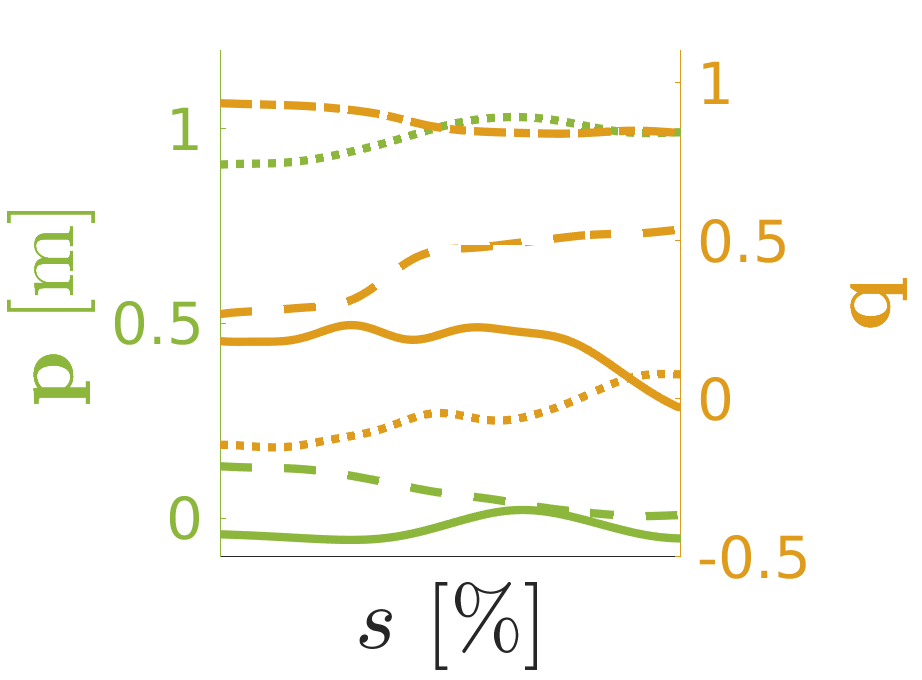} 
		\includegraphics[width=.08\textwidth, trim={0.1cm 2cm 0.1cm 2cm}]{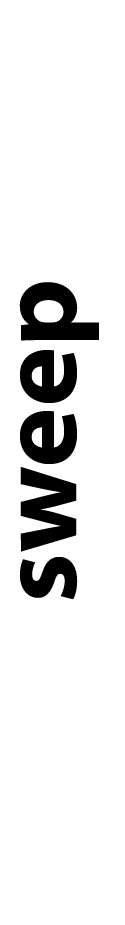}
		\includegraphics[width=.88\textwidth]{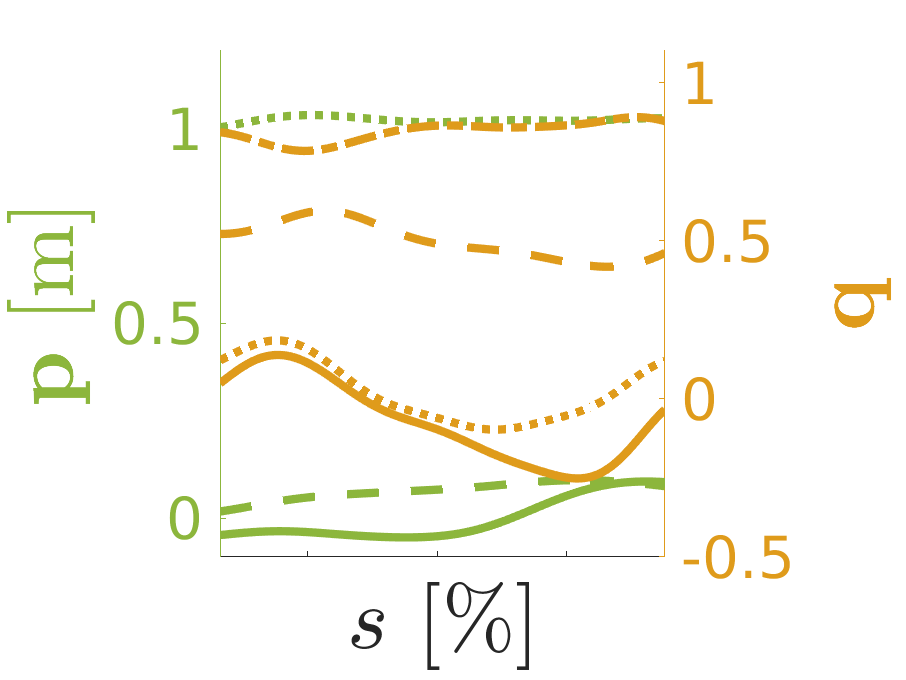}
		\caption{$\mathsf{C}_{\mathsf{approach}}, \mathsf{C}_{\mathsf{sweep}}$}
		\label{subFig:HumanSkills}
	\end{subfigure}
	\begin{subfigure}[b]{0.24\textwidth}
		\includegraphics[width=1.8\textwidth, trim={-6cm -0.2cm 0 0},clip]{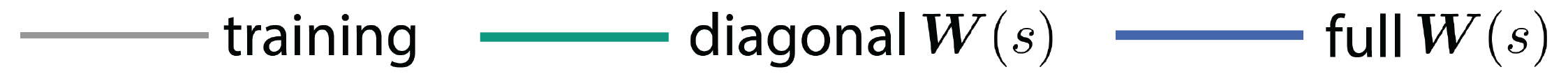}
		\includegraphics[width=\textwidth]{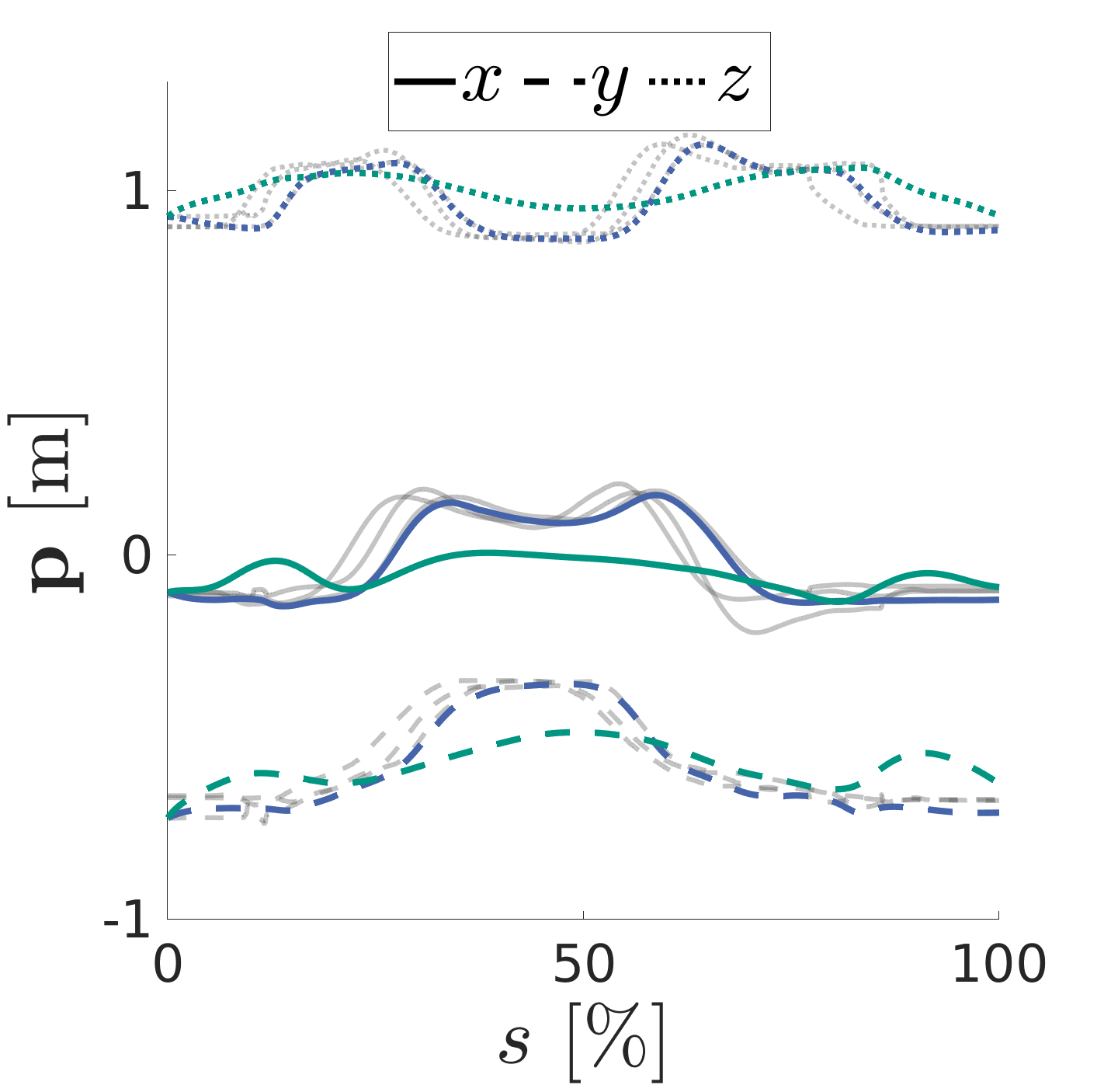}
		\caption{Position (left arm)}
		\label{subFig:HumanPosition}
	\end{subfigure}
	\begin{subfigure}[b]{0.24\textwidth}
		\includegraphics[width=\textwidth]{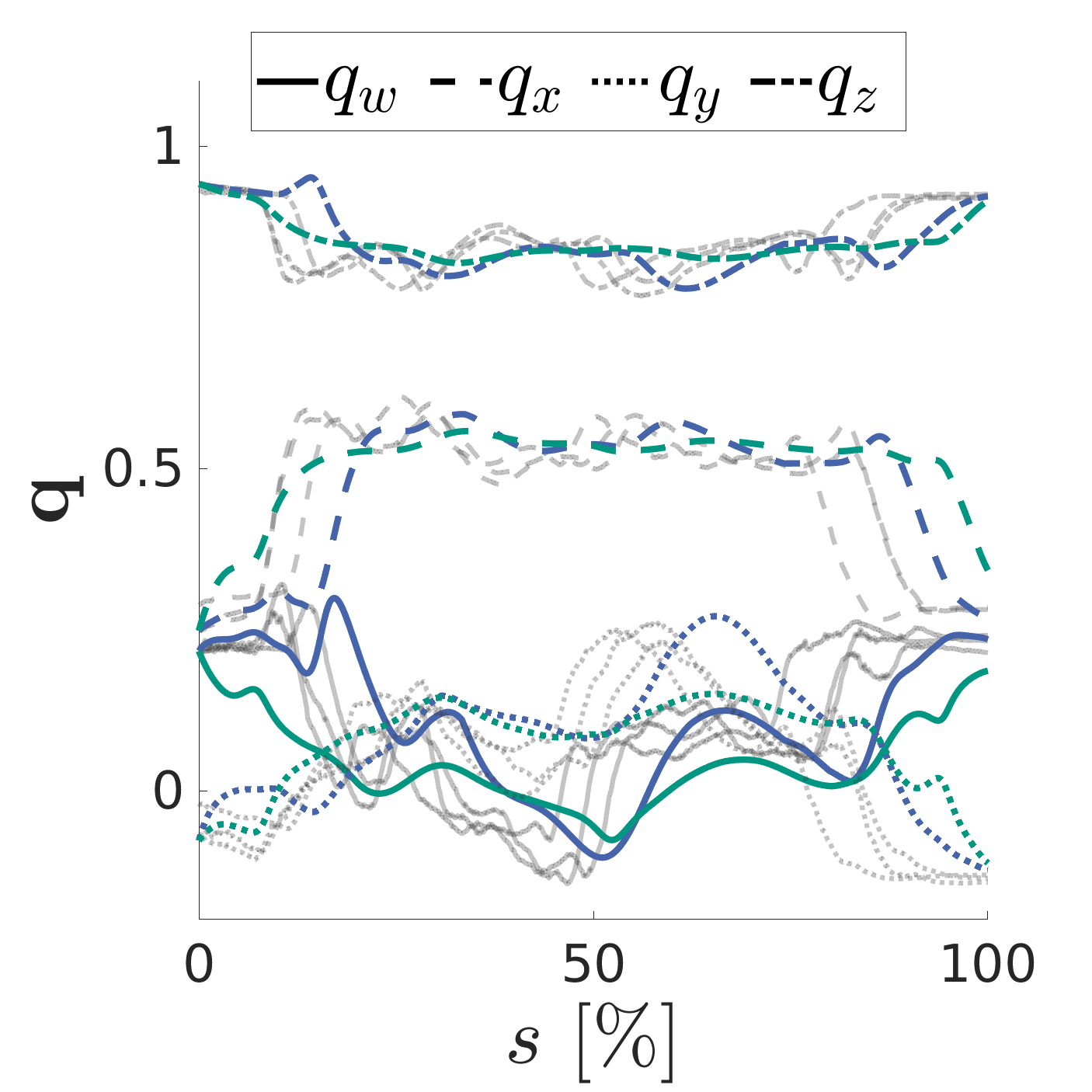}
		\caption{Orientation (right arm)}
		\label{subFig:HumanOrientation}
	\end{subfigure}
	\begin{subfigure}[b]{0.13\textwidth}
		\includegraphics[width=\textwidth]{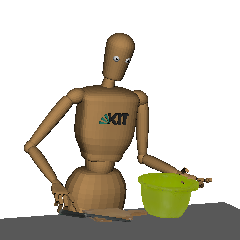}
		\includegraphics[width=\textwidth]{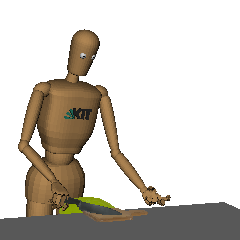}
		\caption{Diag. $\bm{W}(s)$}
		\label{subFig:HumanSnapshotsDiag}
	\end{subfigure}
	\begin{subfigure}[b]{0.13\textwidth}
		\includegraphics[width=\textwidth]{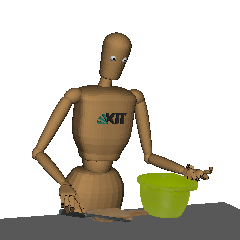}
		\includegraphics[width=\textwidth]{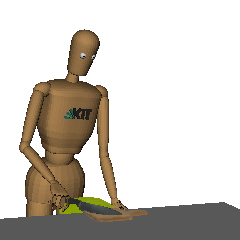}
		\caption{Full $\bm{W}(s)$}
		\label{subFig:HumanSnapshotsFull}
	\end{subfigure}
	\caption{Bimanual sweeping task with a human model. \emph{(a)} $\mathsf{Approach}$ and $\mathsf{sweep}$ VMP skills. \emph{(b)}-\emph{(c)} Demonstrations (\graytraining) and reproductions of the task using diagonal (\kitgreenrepro) and full (\kitbluerepro) weight matrices. \emph{(d)}-\emph{(e)} Snapshots of the reproduction at $s=0.25$ (\emph{top}) and $s=0.5$ (\emph{bottom}).}
	\label{Fig:MMMexperiment}
	\vspace{-0.35cm}
\end{figure*}

We aim at evaluating our approach to sequence and blend skills based on human demonstrations, i.e., on data for which no ground truth is easily available. To do so, we consider a bimanual sweeping task from the KIT motion database~\cite{Mandery16:MotionDatabase, Krebs21:MotionDatabase}, 
in which a human transfers cucumber slices from a cutting board to a bowl. 
At the beginning of the demonstrations, a subject stands in front of a table.
A cutting board on which cucumber slices are placed, is positioned along the edge of the table in front of the human.
The human first grasps a plastic bowl with the left hand and a knife with the right hand using cylindrical power grasps. 
Then, s/he holds the bowl below the table next to the cutting board, and pushes the cucumber slices into the bowl with the knife. Finally, the human places both knife and bowl back. 

For the bimanual sweeping task, we consider the motion of each arm separately. Moreover, we use demonstrations of the aforementioned sweeping task performed by two different subjects. First, three naturally-varying demonstrations of the first subject are used to obtain a skill library. Here, we consider a set of four low-level skills per arm, namely $\{\mathsf{C}^{\text{l}}_{\mathsf{approach}},\mathsf{C}_{\mathsf{hold}}, \mathsf{C}^{\text{l}}_{\mathsf{place}},\mathsf{C}^{\text{l}}_{\mathsf{retreat}}\}$ and $\{\mathsf{C}^{\text{r}}_{\mathsf{approach}},\mathsf{C}_{\mathsf{sweep}}, \mathsf{C}^{\text{r}}_{\mathsf{place}},\mathsf{C}^{\text{r}}_{\mathsf{retreat}}\}$ for the left and right arm, respectively. Each human demonstration is manually segmented into four parts corresponding to the $\mathsf{approach}$, $\mathsf{sweep}$/$\mathsf{hold}$, $\mathsf{place}$, and $\mathsf{retreat}$ skills. 
In this experiment, we use via-points movement primitives (VMP)~\cite{Zhou2019:VMP}, which offer powerful skill representations that are easily adaptable to new starts, goals and via-points after training. 
Therefore, each skill is then represented by a time-dependent VMP trained on the corresponding segments of the demonstrations. The desired control values are the end-effector position and unit-quaternion-based orientation $\left(\begin{smallmatrix}
\bm{p} \\ \bm{q}
\end{smallmatrix}\right)$ given by the mean trajectory retrieved by the VMPs. The desired control values depend on the time $t_s$, i.e., $\hat{\bm{\xi}}(\bm{\psi}_s) \equiv \left(\begin{smallmatrix}
\hat{\bm{p}}(t_s) \\ \hat{\bm{q}}(t_s)
\end{smallmatrix}\right)$. 
All VMPs are executed with the start and goal poses defined by the desired task.
The timing of the VMP skills is defined by the duration of the entire task $T$.
Within our model, every skill trajectory is then evaluated at the evolving time $t = s T$ based on the overall phase variable $s$.
The resulting skills are illustrated by Fig.~\ref{subFig:HumanSkills}.
As for the previous experiments, these skills are considered as black-box solutions, meaning that their representation is not directly known by our model.
We then use three demonstrations provided by a second \emph{different} subject to train two models of our approach (left and right arm separately) with diagonal and full weight matrices. 
Note that these demonstrations include variations, as humans motions naturally vary across executions of the same task.

A simulated kinematic human model, as well as models of the bowl, knife and table, are used for the reproduction phase.
In this case, the model with diagonal weight matrices could not reproduce the task as it was not able to closely fit the demonstrations (see Fig.~\ref{subFig:HumanPosition}-~\ref{subFig:HumanSnapshotsFull}). This is due to the significant differences between the low-level skill trajectories (trained on the first subject) and the demonstrations (provided by the second subject). Notice that such differences also appeared in the pick-and-place experiment. However, as opposed to the sweeping task, the arm trajectories between the pick and place locations did not influence the task success, allowing both diagonal and full weight matrices to be used.
For the bimanual sweeping task, only full weight matrices lead to a successful reproduction by learning correlations between skills. 
Notice that, although two separated models were trained for the left and right arms, the learned full weight functions conserved the timing of the motions, allowing both arms to be synchronized during the reproduction. Also, the training and testing times were similar to the pouring task.


\section{Conclusion}
\label{sec:Conclusion}
We proposed a skill-agnostic formulation to learn to sequence and blend skills using QP-based differentiable optimization layers. This allows us to represent the relative importance of skills as a function of the task progress and to optimize it for a given loss with gradient-based approaches. Our experiments showed that, provided a set of black-box skills and one or few demonstrations of a task, our approach not only learns unknown sequences composed of various types of skills, but also generates smooth motions with seamless, blended transitions. Overall, our diagonal model is advantageous for generalization, while full weight matrices are beneficial when demonstrations must be closely followed.

It is worth noticing that the considered pouring and sweeping tasks are generally difficult to learn with a single model. Instead, our approach decomposes a task by combining several skills, which are easy to train and potentially re-usable across tasks. Moreover, it requires only one or few demonstrations of the complete task, making it less cumbersome to train than trial-and-error-based models. This is a major advantage compared to black-box optimization techniques used in multitask control, although detailed performance comparisons are deferred to future work.
Finally, in contrast to end-to-end methods, our formulation is modular, fast to train, and interpretable as the relative importance of skills is directly embedded in the weight matrices.

Importantly, the performance of our approach highly depends on the capabilities of the given individual skills. Namely, a given task can be reproduced only if the provided skill library contains a set of skills that can be sequenced and combined to do so. Also, our approach generalizes to new object locations under the condition that the corresponding skills successfully adapt to these locations. The dependency of the model parameters to a time-driven phase variable also limits the generalization. This can be overcome by defining the phase variable as a time-independent, perception-based measure of task progress, which we will explore in the future.

One drawback of our approach is that the dimensionality of the optimization variable increases rapidly with the number of different types of skills, i.e., which provide different control variables. To be applied to cases featuring a complex library with many different types of skills, we will extend our approach to handle hierarchies of skills. For instance, high-level skills, e.g., $\mathsf{sweeping}$ cucumber slices to a bowl, may first be learned with our approach as sequences of low-level skills, and then combined in a complex task, e.g., $\mathsf{preparing}$ a salad, with an additional QP-based formulation. 
We will then evaluate our approach in more complex scenarios including, e.g., hierarchies, and soft prioritization of skills.


\bibliographystyle{IEEEtran}

\bibliography{References} 

\begin{thebibliography}{10}
\providecommand{\url}[1]{#1}
\csname url@rmstyle\endcsname
\providecommand{\newblock}{\relax}
\providecommand{\bibinfo}[2]{#2}
\providecommand\BIBentrySTDinterwordspacing{\spaceskip=0pt\relax}
\providecommand\BIBentryALTinterwordstretchfactor{4}
\providecommand\BIBentryALTinterwordspacing{\spaceskip=\fontdimen2\font plus
\BIBentryALTinterwordstretchfactor\fontdimen3\font minus
  \fontdimen4\font\relax}
\providecommand\BIBforeignlanguage[2]{{%
\expandafter\ifx\csname l@#1\endcsname\relax
\typeout{** WARNING: IEEEtran.bst: No hyphenation pattern has been}%
\typeout{** loaded for the language `#1'. Using the pattern for}%
\typeout{** the default language instead.}%
\else
\language=\csname l@#1\endcsname
\fi
#2}}

\bibitem{MussaIvaldi00:MotorPrimitives}
F.~Mussa-Ivaldi and E.~Bizzi, ``Motor learning through the combination of
  primitives,'' \emph{Philos. Trans. R. Soc. Lond., B, Biol. Sci.}, vol. 355,
  pp. 1755--1769, 2000.

\bibitem{Flash05:MotorPrimitives}
T.~Flash and B.~Hochner, ``Motor primitives in vertebrates and invertebrates,''
  \emph{Curr. Opin. Neurobiol.}, vol.~15, no.~6, pp. 660--666, 2005.

\bibitem{Johansson09:GraspPhases}
R.~S. Johansson and J.~R. Flanagan, ``Coding and use of tactile signals from
  the fingertips in object manipulation tasks,'' \emph{Nat. Rev. Neurosci.},
  vol.~10, no.~5, pp. 345--359, 2009.

\bibitem{Nocedal06:OptimizationBook}
J.~Nocedal and S.~J. Wright, \emph{Numerical Optimization}, 2nd~ed.\hskip 1em
  plus 0.5em minus 0.4em\relax Springer, 2006.

\bibitem{Amos17:Optnet}
B.~Amos and J.~Z. Kolter, ``{OptNet}: Differentiable optimization as a layer in
  neural networks,'' in \emph{{ICML}}, 2017.

\bibitem{Agrawal19:DifferentiableOptLayers}
A.~Agrawal, B.~Amos, S.~Barratt, S.~Boyd, S.~Diamond, and J.~Z. Kolter,
  ``Differentiable convex optimization layers,'' in \emph{{NeurIPS}}, 2019.

\bibitem{Manschitz15:LearningSequentialSkills}
S.~Manschitz, J.~Kober, M.~Gienger, and J.~Peters, ``Learning movement
  primitive attractor goals and sequential skills from kinesthetic
  demonstrations,'' \emph{Rob. Auton. Syst.}, vol.~74, pp. 97--107, 2015.

\bibitem{Manschitz15:ConcurrentSequentialSkills}
------, ``Probabilistic progress prediction and sequencing of concurrent
  movement primitives,'' in \emph{{IEEE/RSJ} {IROS}}, 2015, pp. 449--455.

\bibitem{Rozo20:LearningSequencing}
L.~Rozo, M.~Guo, A.~G. Kupcsik, M.~Todescato, P.~Schillinger, M.~Giftthaler,
  M.~Ochs, M.~Spies, N.~Waniek, P.~Kesper, and M.~B\"urger, ``Learning and
  sequencing of object-centric manipulation skills for industrial tasks,'' in
  \emph{{IEEE/RSJ} {IROS}}, 2020, pp. 9072--9079.

\bibitem{Konidaris12:SkillTrees}
G.~Konidaris, S.~Kuindersma, R.~Grupen, and A.~Barto, ``Robot learning from
  demonstration by constructing skill trees,'' \emph{IJRR}, vol.~31, no.~3, pp.
  360--375, 2012.

\bibitem{Stulp12:RLSequencingDMP}
F.~Stulp, E.~A. Theodorou, and S.~Schaal, ``{Reinforcement Learning with
  Sequences of Motion Primitives for Robust Manipulation},'' \emph{{IEEE}
  T-RO}, vol.~28, no.~6, pp. 1360--1370, 2012.

\bibitem{Saveriano19:BlendingDMP}
M.~Saveriano, F.~Franzel, and D.~Lee, ``Merging position and orientation motion
  primitives,'' in \emph{{IEEE} {ICRA}}, 2019, pp. 7041--7047.

\bibitem{Paraschos18:ProMP}
A.~Paraschos, C.~Daniel, J.~Peters, and G.~Neumann, ``{Using probabilistic
  movement primitives in robotics},'' \emph{Auton. Robot.}, vol.~42, no.~3, pp.
  529--551, 2018.

\bibitem{Luksch12:HierarchicalSequenceMPs}
T.~Luksch, M.~Gienger, M.~M\"uhlig, and T.~Yoshiike, ``Adaptive movement
  sequences and predictive decisions based on hierarchical dynamical systems,''
  in \emph{{IEEE/RSJ} {IROS}}, 2012, pp. 2082--2088.

\bibitem{Muehlig14:HierarchicalSequenceMPs}
M.~M\"uhlig, A.~Hayashi, M.~Gienger, S.~Iba, and T.~Yoshiike, ``Receding
  horizon optimization of robot motions generated by hierarchical movement
  primitives,'' in \emph{{IEEE/RSJ} {IROS}}, 2014, pp. 129--135.

\bibitem{Salini11:Sequencing}
J.~Salini, V.~Padois, and P.~Bidaud, ``Synthesis of complex humanoid whole-body
  behavior: a focus on sequencing and tasks transitions,'' in \emph{{IEEE}
  {ICRA}}, 2011, pp. 1283--1290.

\bibitem{Dehio15:SoftPriorities}
N.~Dehio, R.~F. Reinhart, and J.~J. Steil, ``Multiple task optimization with a
  mixture of controllers for motion generation,'' in \emph{{IEEE/RSJ} {IROS}},
  2015, pp. 6416--6421.

\bibitem{Modugno16:CMAESSoftPriorities}
V.~Modugno, U.~Chervet, G.~Oriolo, and S.~Ivaldi, ``Learning soft task
  priorities for safe control of humanoid robots with constrained stochastic
  optimization,'' in \emph{{IEEE/RAS} Humanoids}, 2016, pp. 101--108.

\bibitem{Su18:BOTaskPriorities}
Y.~Su, Y.~Wang, and A.~Kheddar, ``Sample-efficient learning of soft task
  priorities through {B}ayesian optimization,'' in \emph{{IEEE/RAS} Humanoids},
  2018, pp. 1--6.

\bibitem{Li20:BOSoftPriorities}
J.~Li, Y.~Zhu, L.~Huo, and Y.~Chen, ``Sample-efficient learning of soft
  priorities for safe control with constrained {B}ayesian optimization,'' in
  \emph{{IEEE} {IRC}}, 2020, pp. 406--407.

\bibitem{Bouyarmane16:WeightPriorizedControl}
K.~Bouyarmane and A.~Kheddar, ``On weight-prioritized multitask control of
  humanoid robots,'' \emph{{IEEE} Trans. Autom. Control}, vol.~63, no.~6, pp.
  1542--1557, 2016.

\bibitem{Collette08:QP}
C.~Collette, A.~Micaelli, C.~Andriot, and P.~Lemerle, ``Robust balance
  optimization control of humanoid robots with multiple non coplanar grasps and
  frictional contacts,'' in \emph{{IEEE} {ICRA}}, 2008, pp. 3187--3193.

\bibitem{Kuhn51:KKTconditions}
H.~W. Kuhn and A.~W. Tucker, ``{Nonlinear programming},'' in \emph{Berkeley
  Symp. on Mathematical Statistics and Probability}, 1951, pp. 481--492.

\bibitem{Englert17:InverseKKT}
P.~Englert, N.~A. Vien, and M.~Toussaint, ``Inverse {KKT}: Learning cost
  functions of manipulation tasks from demonstrations,'' \emph{IJRR}, vol.~36,
  no. 13-14, pp. 1474--1488, 2017.

\bibitem{Pirotta16:InverseRL}
M.~Pirotta and M.~Restelli, ``Inverse reinforcement learning through policy
  gradient minimization,'' in \emph{{AAAI}}, 2016, pp. 1993--1999.

\bibitem{Gribovskaya11:DS}
E.~Gribovskaya, S.~M. Khansari-Zadeh, and A.~Billard, ``Learning non-linear
  multivariate dynamics of motion in robotic manipulators,'' \emph{IJRR},
  vol.~30, no.~1, pp. 80--117, 2011.

\bibitem{Zhou2019:VMP}
Y.~Zhou, J.~Gao, and T.~Asfour, ``Learning via-point movement primitives with
  inter- and extrapolation capabilities,'' in \emph{{IEEE/RSJ} {IROS}}, 2019,
  pp. 4301--4308.

\bibitem{Khansari14:DS}
S.~M. Khansari-Zadeh and A.~Billard, ``Learning control {L}yapunov function to
  ensure stability of dynamical system-based robot reaching motions,''
  \emph{Rob. Auton. Syst.}, vol.~62, no.~6, pp. 752--765, 2014.

\bibitem{Bhatia07:PDmatrices}
R.~Bhatia, \emph{Positive Definite Matrices}.\hskip 1em plus 0.5em minus
  0.4em\relax Princeton University Press, 2007.

\bibitem{Asfour19:Armar6}
T.~Asfour, M.~W\"achter, L.~Kaul, S.~Rader, P.~Weiner, S.~Ottenhaus, R.~Grimm,
  Y.~Zhou, M.~Grotz, and F.~Paus, ``{ARMAR}-6: A high-performance humanoid for
  human-robot collaboration in real world scenarios,'' \emph{{IEEE} RAM},
  vol.~26, no.~4, pp. 108--121, 2019.

\bibitem{Mandery16:MotionDatabase}
C.~Mandery, O.~Terlemez, M.~Do, N.~Vahrenkamp, and T.~Asfour, ``Unifying
  representations and large-scale whole-body motion databases for studying
  human motion,'' \emph{{IEEE} T-RO}, vol.~32, no.~4, pp. 796--809, 2016.

\bibitem{Krebs21:MotionDatabase}
F.~Krebs, A.~Meixner, I.~Patzer, and T.~Asfour, ``The {KIT} bimanual
  manipulation dataset,'' in \emph{{IEEE/RAS} Humanoids}, 2020-2021.

\end{thebibliography}

\end{document}